\documentclass[11pt]{article}
\usepackage{amsmath}
\usepackage{subcaption}
\usepackage[preprint]{acl}
\usepackage[most]{tcolorbox}
\usepackage{times}
\usepackage{latexsym}
\usepackage{enumitem}
\usepackage[T1]{fontenc}
\usepackage{hyperref}

\usepackage[utf8]{inputenc}

\usepackage{microtype}

\usepackage{inconsolata}

\usepackage{graphicx}
\usepackage{diagbox}
\definecolor{papergreen}{HTML}{008000} 
\definecolor{paperred}{HTML}{D00000}   
\definecolor{seedblue}{HTML}{2E5AA8}

\title{MathDoc: Benchmarking Structured Extraction and Active \\Refusal on Noisy Mathematics Exam Papers}

\author{
   \textbf{Chenyue Zhou\textsuperscript{5}},
   \textbf{Jiayi Tuo\textsuperscript{6}},
   \textbf{Shitong Qin\textsuperscript{2}},
   \textbf{Wei Dai\textsuperscript{3}},
   \textbf{Mingxuan Wang\textsuperscript{1}},\\
   \textbf{Ziwei Zhao\textsuperscript{1}}
   \textbf{Duoyang Li\textsuperscript{1}},
   \textbf{Shiyang Su\textsuperscript{1}},
   \textbf{Yanxi Lu\textsuperscript{1}},
   \textbf{Yanbiao Ma\textsuperscript{1,3,4}\thanks{*Corresponding author}} \\ 
   \textsuperscript{1}Gaoling School of Artificial Intelligence, Renmin University of China Beijing, China \\
   \textsuperscript{2}Gaotu Techedu Inc.
   \textsuperscript{3}Beijing Key Laboratory of Research on Large Models and Intelligent Governance \\
   \textsuperscript{4}Engineering Research Center of Next-Generation Intelligent Search and Recommendation, MOE \\
   \textsuperscript{5}Nanjing University of Aeronautics and Astronautics \\
   \textsuperscript{6}University of Science and Technology of China \\
}

\begin{document}
\maketitle
\begin{abstract}
The automated extraction of structured questions from paper-based mathematics exams is fundamental to intelligent education, yet remains challenging in real-world settings due to severe visual noise. Existing benchmarks mainly focus on clean documents or generic layout analysis, overlooking both the structural integrity of mathematical problems and the ability of models to actively reject incomplete inputs. We introduce MathDoc, the first benchmark for document-level information extraction from authentic high school mathematics exam papers. MathDoc contains \textbf{3,609} carefully curated questions with real-world artifacts and explicitly includes unrecognizable samples to evaluate active refusal behavior. We propose a multi-dimensional evaluation framework covering stem accuracy, visual similarity, and refusal capability. Experiments on SOTA MLLMs, including Qwen3-VL and Gemini-2.5-Pro, show that although end-to-end models achieve strong extraction performance, they consistently fail to refuse illegible inputs, instead producing confident but invalid outputs. These results highlight a critical gap in current MLLMs and establish MathDoc as a benchmark for assessing model reliability under degraded document conditions. Our project repository is available at \href{https://github.com/winnk123/papers/tree/master}{GitHub repository}
\end{abstract}

\section{Introduction}
In modern educational~\citep{hariyanto2025artificial} scenarios, the complete and accurate extraction of questions from paper-based mathematics examinations into structured databases is a prerequisite for intelligent test assembly and personalized instruction~\citep{yan2025mathagent}. Furthermore, constructing high-quality structured question banks provides high-value corpus support for training the mathematical reasoning~\citep{liu2025cmm,chen2025advancing, chen2025mint,liu2025deepseek} capabilities of Large Language Models (LLMs). However, information extraction from examination documents in real-world scenarios faces several critical challenges:

\begin{itemize}
\addtolength{\leftskip}{-1.8em}  
\item \textbf{Noise Interference:} Handwritten solutions and grading marks frequently overlap with printed text, complicating the extraction of question stems.

\item \textbf{Figure Extraction:} Flexible layouts complicate text-diagram association, requiring simultaneous localization and contextual alignment.

\item \textbf{Invalid Information Rejection:} Real-world artifacts~\citep{wang2025wilddoc} like creases or occlusions frequently truncate question content. A key challenge lies in enabling models to actively identify and reject such incomplete information.
\end{itemize}

\begin{figure}[t]
  \centering
  \vskip -0.1in
  \includegraphics[width=1\linewidth]{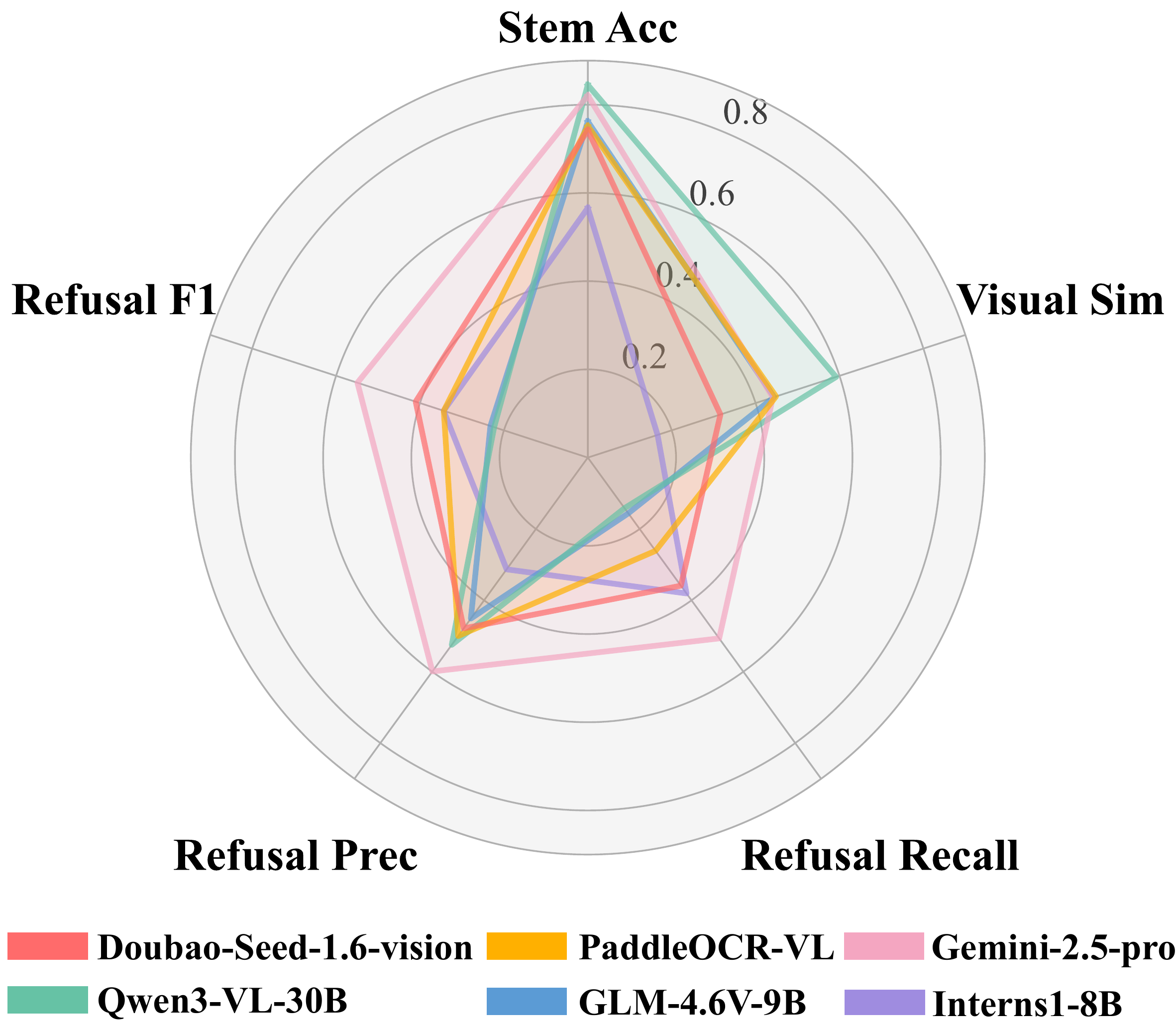}
  \vskip -0.05in
  \caption{Performance comparison of six Multimodal Large Language Models (MLLMs) across five metrics (Stem Accuracy, Visual Similarity, Refusal F1/Precision/Recall) in choice questions.} 
  \label{fig:radar_results}
\vskip -0.1in
\end{figure}

While the rapid evolution of MLLMs~\citep{zhu2025internvl3,rasheed2024glamm,li2024llava,meng2024deepstack,bai2025qwen2,hu2024mplug} and the establishment of benchmarks like OCRBench v2~\citep{fu2024ocrbench}, OmniDocbench~\citep{ouyang2025omnidocbench} have jointly advanced general document information extraction, current evaluations remain inadequate for authentic mathematics documents. Data-wise, existing benchmarks prioritize clean documents with standardized layouts, overlooking real-world noise. Task-wise, they operate on the assumption of input completeness, evaluating only extraction correctness while neglecting the critical capability to actively reject incomplete inputs.

To bridge this gap, we introduce MathDoc, a large-scale benchmark of 3,609 high school mathematics questions captured in real educational settings. 
As is shown in Figure~\ref{fig:radar_results}, the radar chart illustrates the performance of selected models across different evaluation metrics in choice questions. We conducted comprehensive experiments on both open-source models (\textcolor{papergreen}{Qwen3-VL~\citep{bai2025qwen3vltechnicalreport}, InternS1~\citep{bai2025interns1scientificmultimodalfoundation}, GLM-4.6V, DeepSeek-OCR~\citep{wei2025deepseek}, MinerU2.5~\citep{niu2025mineru2}, PaddleOCR-VL~\citep{cui2025paddleocr}}) and closed-source models (\textcolor{paperred}{Gemini-2.5-Pro~\cite{comanici2025gemini}, GPT-4o~\citep{hurst2024gpt}, Doubao-Seed-1.6-vision}). Our experiments reveal that while SOTA models demonstrate impressive accuracy in extracting content from recognizable samples, they universally struggle with active refusal when facing incomplete or occluded inputs. Instead of acknowledging uncertainty, these models predominantly exhibit a tendency towards speculative completion and forced transcription, highlighting a critical gap in the reliability of current MLLMs for real-world applications.

In summary, the contributions of this paper are summarized as follows:
\begin{itemize}
    \item We introduce \textbf{MathDoc}, the first benchmark for document-level information extraction from real-world high school mathematics exam papers. The benchmark consists of \textbf{3,609 high-quality math problems} and establishes a novel \textbf{multi-dimensional evaluation protocol} to comprehensively assess a model’s ability in problem statement identification under noisy document conditions, figure extraction, and answer refusal.
    
    \item We conduct a systematic evaluation of \textbf{SOTA MLLMs}, covering both open-source and close-source models, providing a comprehensive comparison of their performance.
    
    \item We conduct a comprehensive analysis to determine the decision boundaries of active refusal in current MLLMs, quantifying the threshold of information loss to trigger refusal. Furthermore, we propose a simple yet effective method to enhance the refusal capability.
\end{itemize}

\section{Related Works}
\subsection{General Document Information Extraction Benchmarks}
Existing benchmarks~\citep{yang2025cc,heakl2025kitab,liu2024ocrbench} for document information extraction primarily focus on layout analysis~\citep{zhong2019publaynet,li2020docbank,zhong2020image} and visual question answering~\citep{zhou2025dogr,ma2024mmlongbench,mathew2021docvqa,kim2019textbook}, treating documents as collections of generic layout elements~\citep{jaume2019funsd} or OCR tokens. However, they fail to assess unit-level structural integrity, which is critical for mathematical parsing. Specifically, current metrics overlook the cohesive extraction of ``question units" (integrated stems, options, and figures), leaving a significant gap in evaluating question-level structural parsing~\citep{blecher2023nougat} capabilities.

\subsection{Mathematics Specific Document Information Extraction Benchmarks}
Exploration within the specific domain of mathematics document information extraction remains relatively scarce. Existing research falls into two primary categories: one focuses on formula extraction~\citep{horn2025benchmarking,bai2025complex,li2020improving}; the other~\citep{lu2023mathvista,zhang2024mathverse,feng2025mathreal,ye2025logicocr}, comprising the majority of recent studies, focuses on mathematical problem-solving capabilities under idealized visual inputs, rather than low-level document restoration capabilities. Crucially, both categories overlook complex document inputs from authentic educational scenarios, resulting in a lack of effective evaluation regarding model extraction performance under noise interference.

\begin{figure*}[t] \centering \includegraphics[width=0.97\linewidth]{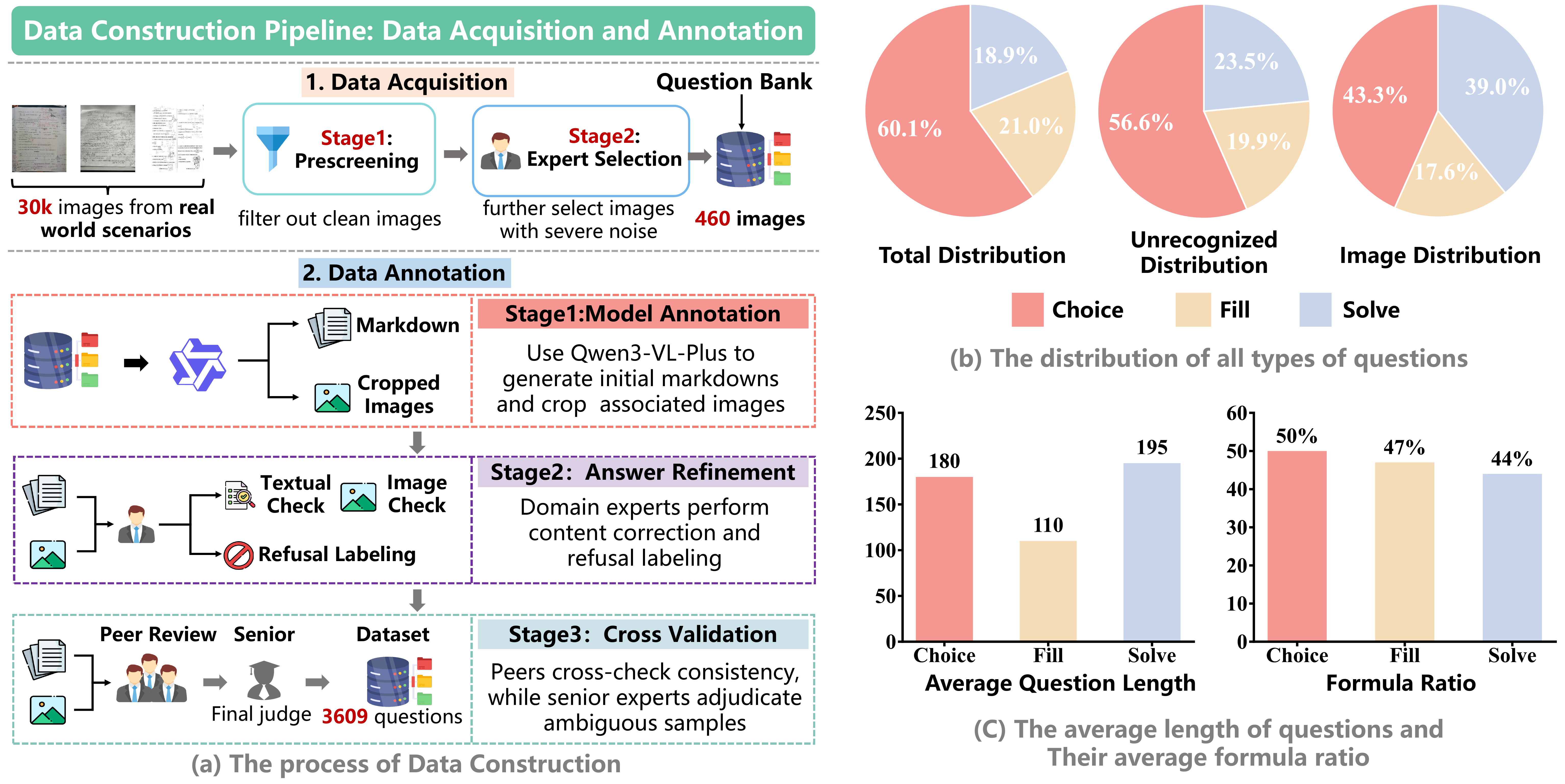}
\vskip -0.05in
\caption{(a) The process of Data Construction. (b) The distribution of all types of questions. (c) The average length of questions and their average formula ratio.} 
\label{fig:dataconstruction} 
\vskip -0.1in
\end{figure*}

\section{Dataset Construction}
\subsection{Data Acquisition}
We collected approximately \textbf{30,000} raw exam images from real-world high school mathematics scenarios, capturing diverse capture devices and physical conditions. To construct a benchmark that rigorously evaluates model robustness, as is illustrated in Figure~\ref{fig:dataconstruction}(a) we curated \textbf{460} highly challenging images via a two-stage pipeline:

\textbf{Layout Pre-screening:} 
We explicitly filter out clean, simple samples to focus on complex geometries. We retain only those featuring multi-column layouts, dense text-figure interleaving to simulate authentic analysis challenges.

\textbf{Expert Selection:} 
Human experts further identify samples with severe OCR noise, specifically targeting: 
(1) heavy handwriting occlusions and corrections in critical text regions; 
(2) physical obstructions (e.g., stains, wrinkles); 
and (3) compromised document integrity, such as incomplete questions or folded pages.

\subsection{Data Annotation}
To balance high-quality ground truth with annotation efficiency, as is illustrated in Figure~\ref{fig:dataconstruction}(a) we established a standardized three-stage pipeline:

\textbf{Model-Assisted Pre-annotation:} 
Leveraging the SOTA model Qwen3-VL-Plus, we generate initial structural Markdown sequences and automatically crop associated geometric figures. This provides a high-quality baseline, significantly reducing the manual workload of starting from scratch.

\textbf{Answer Refinement:} 
Domain experts perform character-level verification under two strict protocols: (1) \textbf{Content Correction}, which involves rectifying textual errors and ensuring precise figure cropping; and (2) \textbf{Refusal Labeling}, where samples rendered illegible by noise are strictly assigned type-specific placeholders (\texttt{[Unrecognizable Choice]}). This establishes a basis for refusal benchmarking.

\textbf{Cross-Validation:} 
To eliminate subjective bias, particularly in defining the boundary between ``recognizable" and ``unrecognizable," we implement a two-tier quality control mechanism: peer cross-checks (focusing on \LaTeX{} syntax and refusal consistency), and a final adjudication by senior experts for ambiguous borderline samples.


\subsection{Dataset Properties}
As is illustrated in Figure~\ref{fig:dataconstruction}(b), the charts present the proportion of Choice, Fill, and Solve questions for overall benchmark and its specific subsets. The first chart represents the overall distribution, which consists of \textbf{2,169} Choice, \textbf{758} Fill, and \textbf{682} Solve questions. The second chart shows the distribution within the subset of questions containing figures, comprising \textbf{162} Choice, \textbf{66} Fill, and \textbf{146} Solve.The third chart depicts the distribution within the ``Unrecognizable'' subset, which includes \textbf{381} Choice, \textbf{134} Fill, and \textbf{158} Solve.

Regarding content details in Figure~\ref{fig:dataconstruction}(c), the average character length per question ranges from \textbf{110 to 195}, with Solve questions being the longest. Furthermore, mathematical formulas constitute a substantial portion of the text (ranging from 44\% to 50\%), requiring models to possess robust capabilities in complex formulation recognition.


\section{Benchmark Construction}
\subsection{Task Definition}
We define this task as a document structured extraction problem. Given an input image of a mathematics exam paper $I$, the goal is to parse it and generate a structured set of problems:

\begin{equation}
\mathcal{S} = \{q_1, q_2, \dots, q_N\}.
\end{equation}

This task requires the model to simultaneously perform three sub-tasks: structured text extraction, figure extraction, and intelligent rejection. 

\noindent\textbf{Structured Text Extraction:} The extracted problem text $T_i$ must not only match the content in the document exactly, but also preserve the reading order consistent with human perception (top-to-bottom, left-to-right).

\noindent\textbf{Figure Extraction:} For problems $q_i$ containing geometric figures, the model is required to predict the bounding box coordinates $B_i$ of the figure, where the corresponding cropped image region must exhibit high semantic similarity with the ground-truth figure.

\noindent\textbf{Intelligent Rejection}: Only problems with complete visual information should be extracted. We define a ground-truth-based completeness function 

\begin{equation}
\Phi(I, q_i) \in \{0, 1\}.
\end{equation}

The expected output of the model $O_i$ is:

\begin{equation}
O_i =
\begin{cases} 
T_i, & \text{if } \Phi(I, q_i) = 1 \\ 
\mathtt{\langle REJECT \rangle}, & \text{if } \Phi(I, q_i) = 0
\end{cases}
\end{equation}

Here, $\Phi = 0$ (ground-truth rejection) is defined by the following two cases:

\begin{itemize}
\addtolength{\leftskip}{-1em}
\item \textbf{Critical Occlusion}: The core region of the problem is completely covered.
\item \textbf{Physical Truncation}: The problem is located at the edge of image, causing semantic interruption.
\end{itemize}

\subsection{Evaluation Pipeline Construction}

\begin{figure*}[t]
  \centering
  \includegraphics[width=0.98\textwidth]{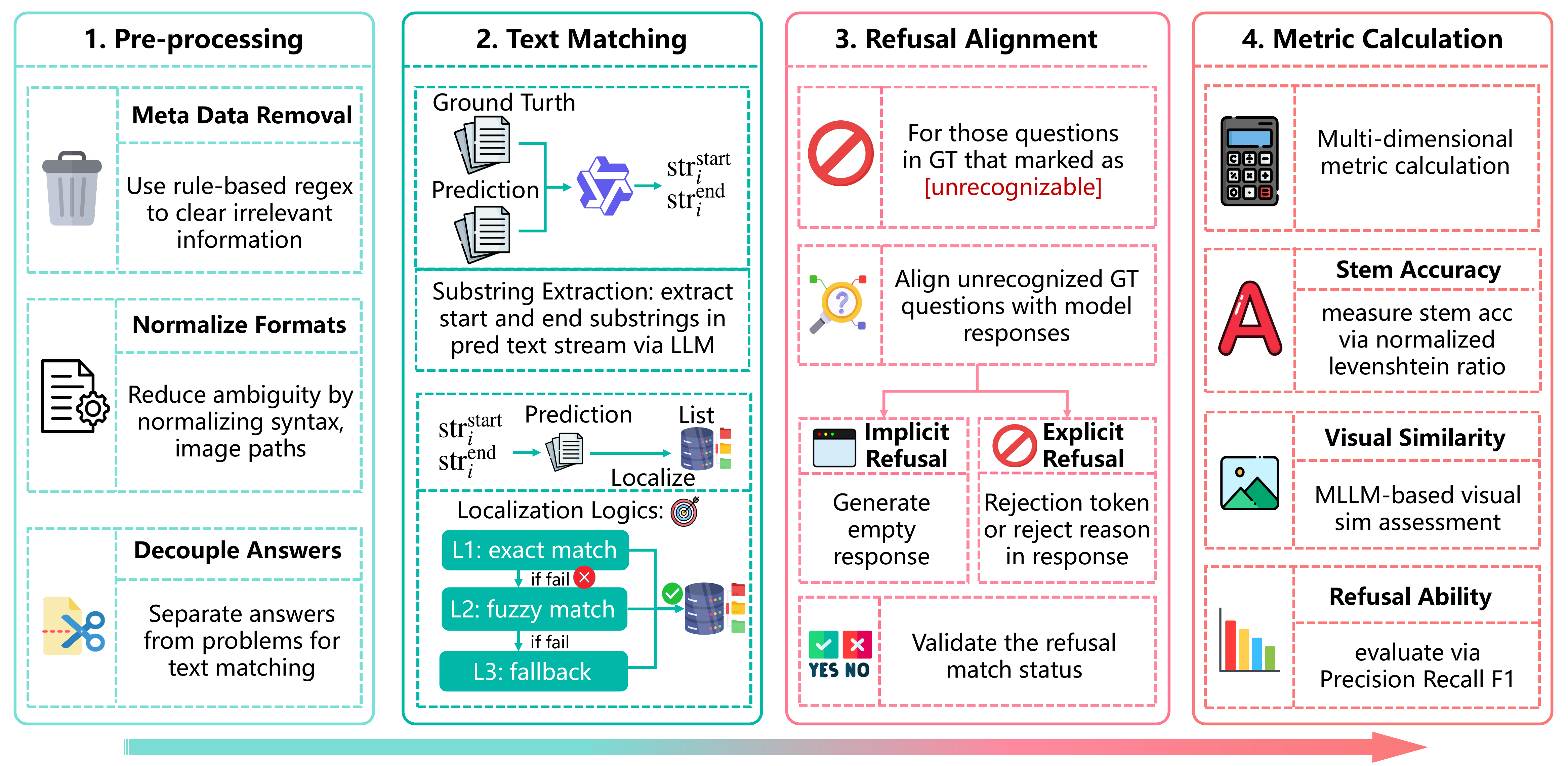}
  \vskip -0.1in
  \caption{The pipeline of evaluation process}
  \label{fig:benchmark pipeline}
\vskip -0.15in
\end{figure*}

To provide a fair and comprehensive evaluation of different models, we propose an end-to-end evaluation pipeline, arranged sequentially in the following steps: pre-processing, text matching, rejection matching, and metric computation (see Figure~\ref{fig:benchmark pipeline}).

\subsubsection{Pre-Processing}
To ensure robust downstream matching, we define a text normalization function $\phi(\cdot)$ that standardizes input sequences through the following operations:

\noindent\textbf{Filtering Non-Semantic Noise:} We eliminate irrelevant metadata (exam instructions, headers) using rule-based regex to reduce interference.

\noindent\textbf{Canonicalizing Formats:} We reduce ambiguity by unifying \LaTeX{} mathematical syntax and converting relative image paths to absolute ones.

\noindent\textbf{Decoupling Reference Answers:} We explicitly separate answers from problem statements to accurately confine the downstream matching scope.

\subsubsection{Text Matching}
This process establishes a fine-grained semantic mapping between valid questions in the Ground Truth (GT) and the model predictions (Pred) by precisely locating question boundaries within the continuous, unstructured prediction stream.

\noindent\textbf{Problem Definition:} Given a GT question set $\mathcal{Q} = \{q_i\}_{i=1}^M$ and a Pred text stream $\mathcal{T}$ of length $N$, the goal is to determine a sequence of boundary indices for precise text matching:
\begin{equation}
    \hat{\Omega} = \{ [\hat{s}_i, \hat{e}_i] \}_{i=1}^M
\end{equation}
Here, $\hat{s}_i$ and $\hat{e}_i$ represent the integer character indices in $\mathcal{T}$,corresponding to the start and end of the $i$-th question. These indices must satisfy the strict sequential constraint:
\begin{equation}
    1 \le \hat{s}_1 < \hat{e}_1 \le \hat{s}_2 < \dots < \hat{e}_M \le N
\end{equation}

\noindent\textbf{Boundary Indices Localization:}
To achieve robust alignment, we propose a hierarchical strategy that synergizes semantic understanding with deterministic search. The process operates in two sequential phases: (1) Substring Extraction, extracting text segments based on semantic correspondence; and (2) Index Localization, determining their precise integer indices.

\begin{itemize}
\addtolength{\leftskip}{-0.1em}
\item[(1)] \textbf{Substring Extraction:}
This phase utilizes the LLM to identify the specific textual segments that demark the boundaries of each question $q_i$ within the noisy prediction stream $\mathcal{T}$. 

\begin{equation} (\text{str}^{\text{start}}_i, \text{str}^{\text{end}}_i) = \operatorname{LLM}(q_i \mid \mathcal{T},\mathcal{Q}) \end{equation}

By jointly processing the ground-truth question set $\mathcal{Q}$ and the predicted text stream $\mathcal{T}$, the model locates the start ($\mathrm{str}_i^{\text{start}}$) and end boundary ($\mathrm{str}_i^{\text{end}}$) of each question $q_i$ within the noisy stream $\mathcal{T}$.

\item[(2)] \textbf{Index Localization:}
Let $k$ denote a candidate start index in the predicted text stream $\mathcal{T}$, used to locate either start anchor $\hat{s}_i$ or end anchor $\hat{e}_i$ of question $q_i$, with the optimal index determined via the following hierarchical search:

\begin{equation}
\hat{k} =
\begin{cases}
\mathrm{find}(\text{str} \subset \mathcal{T}), & \text{L1},\\[1pt]

\displaystyle
\arg\max_{k} \;\mathrm{sim}(\text{str}, \mathcal{T}_k), 
& \text{L2},\\[1pt]

k_{\mathrm{fall}}, & \text{L3}
\end{cases}
\end{equation}
\end{itemize}

\noindent\textbf{Localization Logic:} 
The search logic follows three distinct levels:
\begin{itemize}
\addtolength{\leftskip}{-1em}
\item Level 1 (Exact Match) Scans the text stream to locate the exact literal occurrence of the anchor substring in pred text streams.

\item Level 2 (Fuzzy Match) If exact match fails, a sliding window (8--15 characters) traverses the text. The position with the highest normalized Levenshtein similarity is selected, provided the score exceeds the threshold $\tau = 0.8$.

\item Level 3 (Index Fallback) If both matches fail, the algorithm defaults to the last verified position to maintain continuity. Specifically, it assigns the previous question's end index to a missing start anchor ($\hat{s}_i \leftarrow \hat{e}_{i-1}$) or collapses a missing end anchor to the current start index ($\hat{e}_i \leftarrow \hat{s}_i$).
\end{itemize}

\subsubsection{Refusal Alignment}
For ground truth entries marked as noise (``[Unrecognizable]"), text matching is inapplicable. Instead, we employ a refusal alignment strategy:
\begin{itemize}
\addtolength{\leftskip}{-1em}
\item \textbf{Implicit Refusal:} The model yields an empty output ($\hat{y}_i = \emptyset$) for a certain question.

\item \textbf{Explicit Refusal:} The model generates predefined rejection tokens (e.g., ``[Unrecognizable]") or reasons for refusal to answer, achieving a direct label match with the ground truth.
\end{itemize}

\begin{figure*}[t]
  \centering
  \includegraphics[width=\textwidth]{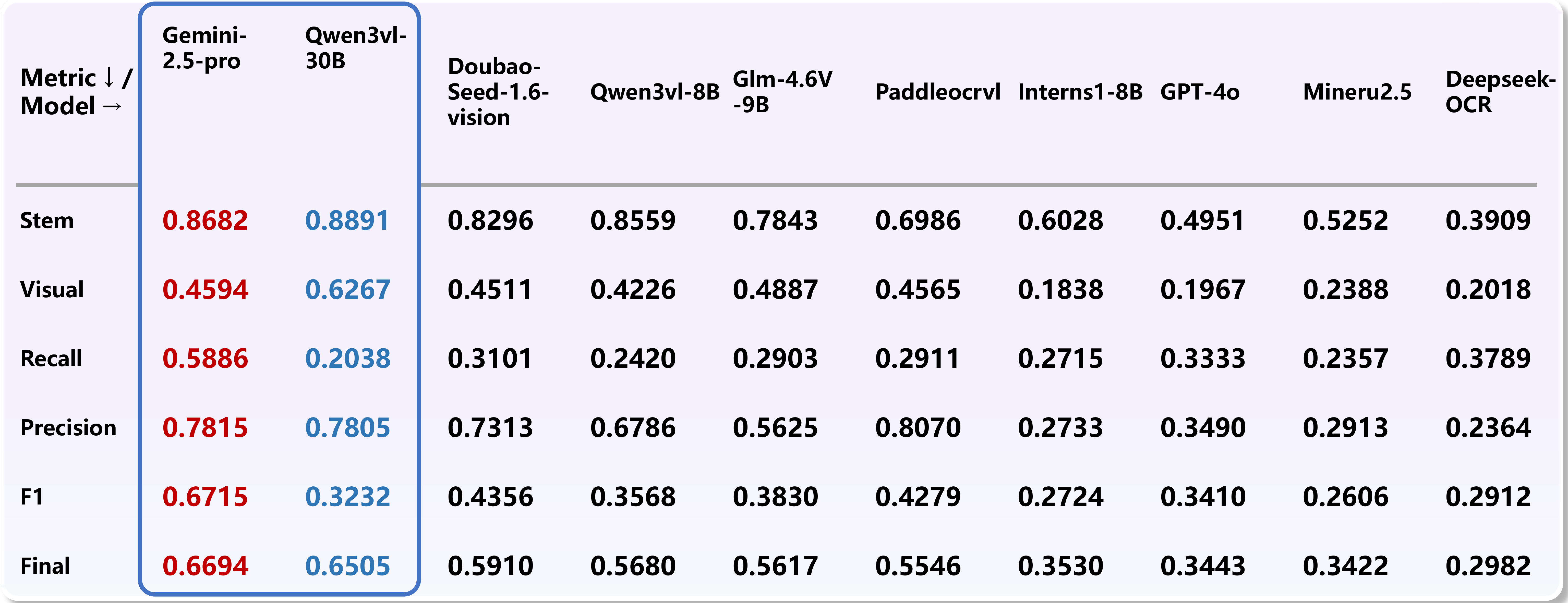}
  \caption{Model performance on Solve, sorted in descending order according to the Final metric (left to right).}
  \label{fig:solve performence}
\end{figure*}

\subsubsection{Evaluation Metrics}
We evaluate the model's performance across three dimensions using the following metrics.

\noindent\textbf{Stem Accuracy:} We utilize the Normalized Levenshtein Ratio to quantify the text extraction similarity across all $M$ questions:

\begin{equation}
\small
S_{\text{text}} = \frac{1}{M} \sum_{i=1}^{M} \left( 1 - \frac{\operatorname{Lev}(\phi(T^{\text{Pred}}_i), \phi(Q^{\text{GT}}_i))}{\max(|\phi(T^{\text{Pred}}_i)|, |\phi(Q^{\text{GT}}_i)|)} \right) \end{equation}

\noindent\textbf{Visual Similarity:} We employ a MLLM (Qwen3VL-Plus) as a Semantic Judge~\citep{chen2024humans} $\mathcal{J}$ to assess the visual similarity between the extracted image $I^{\text{Pred}}_i$ and the ground truth $I^{\text{GT}}_i$:
\begin{equation}S_{\text{img}} = \frac{1}{M} \sum_{i=1}^{M} \mathcal{J}(I^{\text{Pred}}_i, I^{\text{GT}}_i)\end{equation}
where the function $\mathcal{J}$ outputs a semantic similarity score in the range $[0, 1]$

\noindent\textbf{Refusal Capabilities:} We formulate the identification of incomplete questions as a binary classification task, treating Refusal as the positive class. The model's performance is evaluated using standard Precision, Recall, and F1-Score.

\noindent\textbf{Final Performance Score:} 
The final performance score aggregates text, image, and refusal metrics. Specifically, the refusal dimension is composed of Precision, Recall, and F1-score, each contributing equally to the dimension's total weight:
\begin{equation}
\small
S_{\text{final}} =
\frac{1}{3} S_{\text{text}}
+ \frac{1}{3} S_{\text{img}}
+ \frac{1}{9}
\left(
P_{\text{ref}} + R_{\text{ref}} + F1_{\text{ref}}
\right)
\end{equation}

\section{Experiments}
\begin{figure*}[t]
\vskip -0.1in
  \centering
  \includegraphics[width=\textwidth]{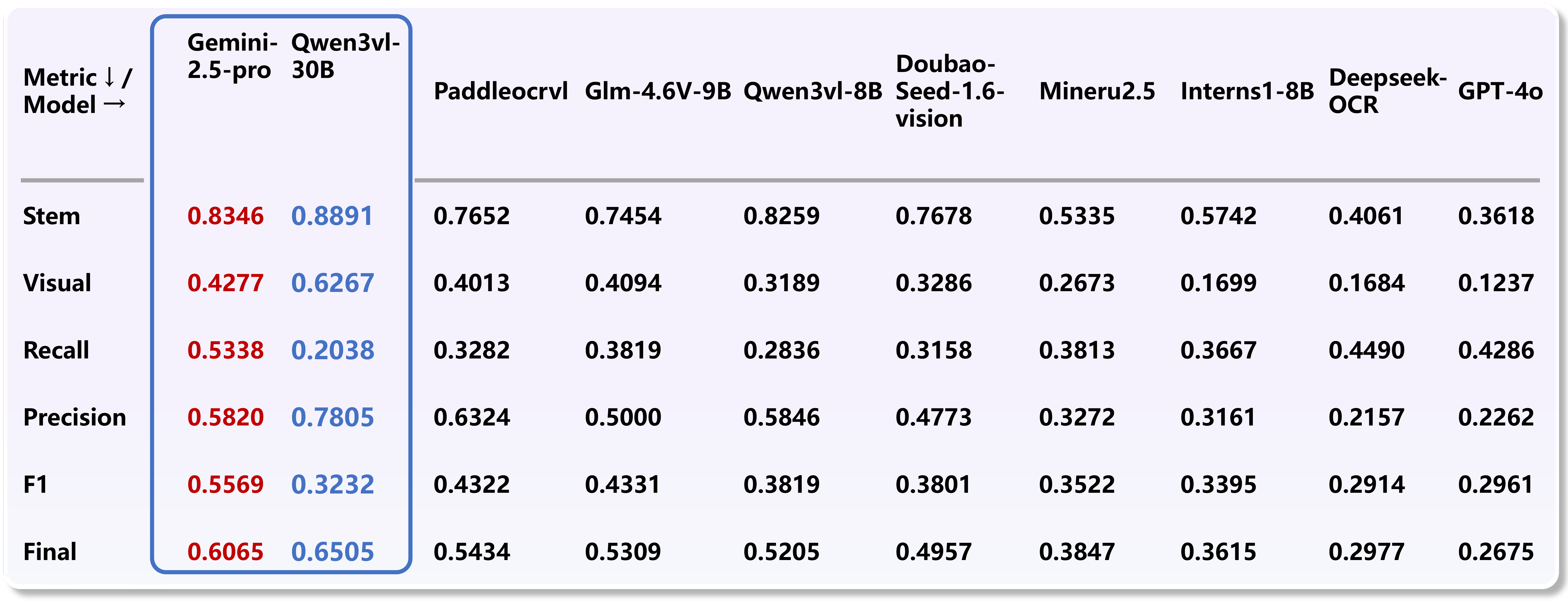}
  \caption{Model performance on Fill, sorted in descending order according to the Final metric (left to right).}
  \label{fig:fill performance}
\end{figure*}

In this section, we benchmark SOTA MLLMs on the \textbf{MathDoc} dataset, presenting a comprehensive quantitative evaluation followed by an in-depth analysis. The results are shown in Figure~\ref{fig:solve performence},~\ref{fig:choice performance} and~\ref{fig:fill performance}. Furthermore, we design two targeted experiments to investigate the underlying mechanisms of the models' refusal capabilities. The results are shown in Figure~\ref{fig:refusual} and~\ref{fig:exper1}.

\subsection{Main Result Analysis}
\begin{figure*}[t]
  \centering
  \includegraphics[width=\textwidth]{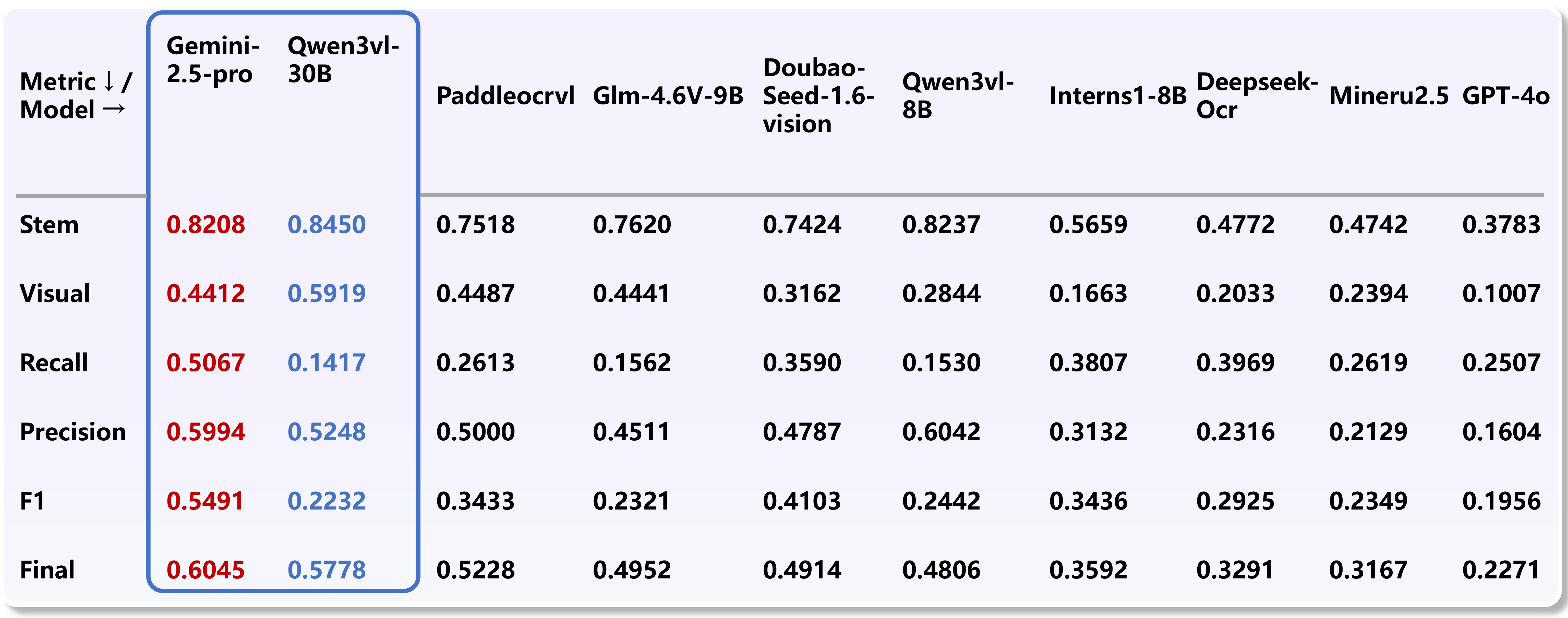}
  \caption{ Model performance on Choice, sorted in descending order according to the Final metric (left to right).}
  \label{fig:choice performance}
\end{figure*}

\textbf{Analysis 1:}
As shown in Figure~\ref{fig:solve performence}, Qwen3-VL-30B achieves state-of-the-art extraction performance ($S_{\text{text}}=0.89$, $S_{\text{img}}=0.63$). However, refusal evaluation reveals a consistent ``Low Recall, High Precision'' pattern across most models (excluding Gemini). For example, despite its strong extraction accuracy, Qwen3-VL-30B attains a refusal recall of only 0.14 (Figure~\ref{fig:choice performance}). \textbf{This imbalance indicates that while models can produce correct refusals when triggered, they rarely initiate refusal proactively.} Instead, they tend to forcibly transcribe fragmented content or speculate missing details. We attribute this behavior to the lack of explicit negative supervision for document incompleteness during pretraining and instruction tuning, resulting in under-trained decision boundaries for completeness-aware rejection.

\textbf{Analysis 2:}
Most models perform better on Problem-Solving tasks than on Choice tasks, primarily due to structural differences. Choices require precise localization of densely packed options, where minor spatial misalignment can cause extraction errors. However, Solve tasks provide richer context, enabling models to leverage broader linguistic priors to mitigate perceptual ambiguities.

\textbf{Analysis 3:}
End-to-End MLLMs generally outperform multi-stage pipeline frameworks (MinerU, PaddleOCR-VL). Pipeline approaches typically suffer from error cascading, where failures in the initial layout analysis phase irreversibly corrupt subsequent text recognition.

\subsection{Investigation of Refusal Mechanisms}
The pervasive low Refusal F1 scores shown in the table necessitate a deeper examination of the underlying refusal behaviors. Consequently, we design targeted experiments to answer two core questions: 

\begin{itemize}
\item Q1: What degree of information loss triggers a model's refusal response? 
\item Q2: What strategies can enhance the model's refusal performance?
\end{itemize}

\subsubsection{Sensitivity to Information Loss}
\textbf{Experimental Setup:} To investigate the specific boundary conditions under which models tend to refuse answering due to missing information, We constructed a evaluation set consisting of 50 independent questions cropped from the original full-page images. To simulate varying degrees of visual incompleteness, we applied a generative erasure technique to all of the images. Specifically, we quantify the extent of information loss based on the textual density within the visual region. We introduce the Information Loss Rate, denoted as $\eta$:
\begin{equation}\eta = \frac{N_{erased}}{N_{total}} \times 100
\end{equation}

Consequently, we define the Refusal Rate $R_{ref}(\eta)$ at a specific occlusion level $\eta$ as:
\begin{equation}
R_{ref}(\eta) = \frac{N_{refused}^{(\eta)}}{N_{set}} \times 100\%
\label{eq:refusal_rate}
\end{equation}
where $N_{refused}^{(\eta)}$ denotes the count of \textbf{explicitly refused} answers under the occlusion rate $\eta$, and $N_{set}$ represents the total number of questions.

\begin{figure}[t] 
\centering \includegraphics[width=1\linewidth]{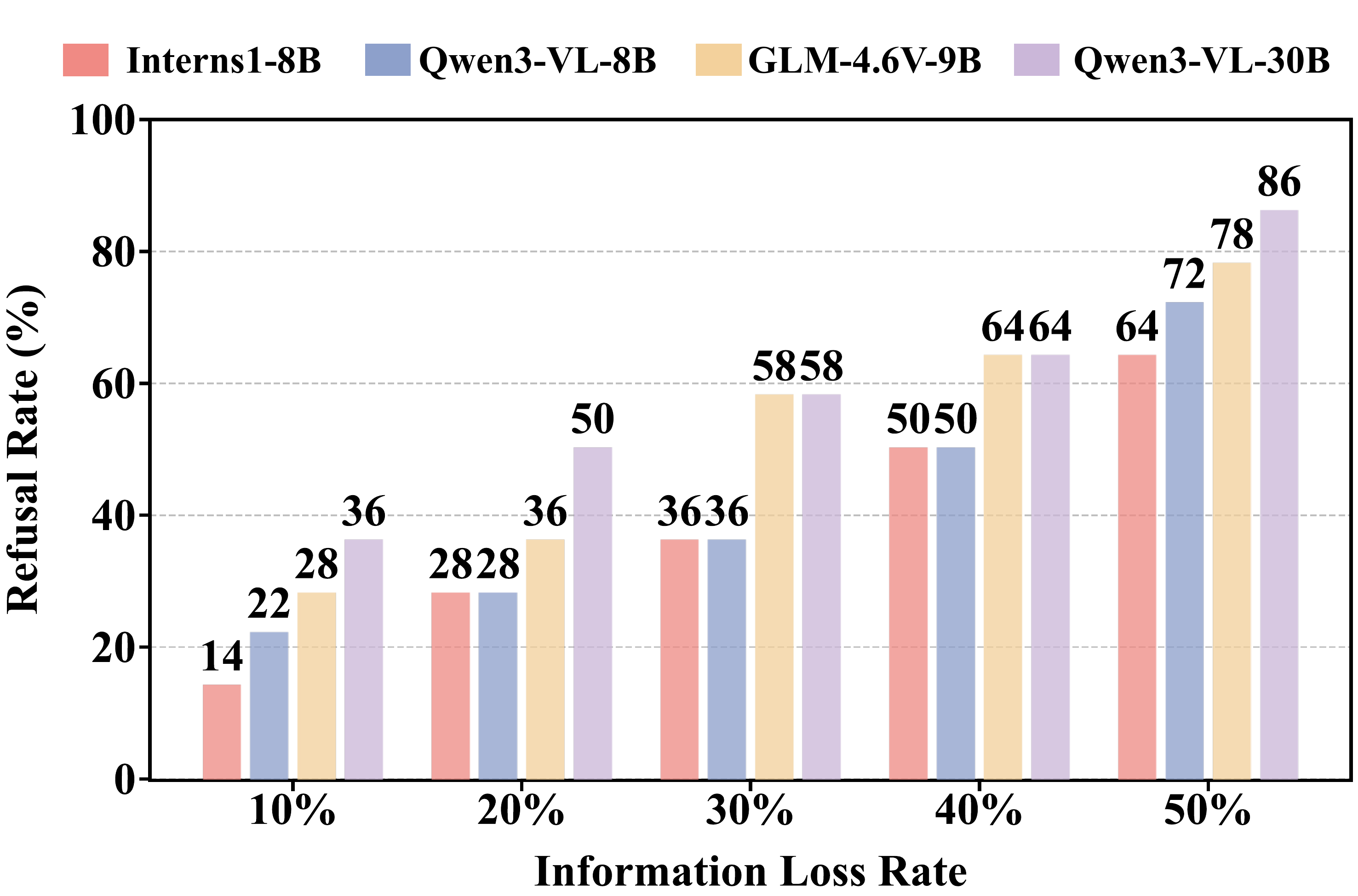} 
\caption{The result of refusal performance.}
\label{fig:refusual} 
\end{figure}

\noindent\textbf{Result Analysis:} As illustrated in Figure~\ref{fig:refusual}, we observe a distinct correlation between model scale and sensitivity to visual occlusion. Taking the 50\% refusal rate as a critical threshold, the larger Qwen3VL-30B exhibits the highest sensitivity, identifying and refusing compromised inputs at a low information loss rate of $\eta=20\%$, followed by GLM at $\eta=30\%$. In contrast, lightweight models like Interns1-8B and Qwen3VL-8B demonstrate lower awareness of information completeness, attempting to answer even until the occlusion reaches $\eta=40\%$. This indicates that larger models possess a \textbf{superior perception of visual integrity}, effectively detecting missing contexts to \textbf{ensure response reliability}, whereas smaller models are less sensitive to information loss, tending to \textbf{process the incomplete visual inputs indiscriminately}.

\subsubsection{Exploration on Strategies for Refusal Enhancement}
In this subsection, we investigate methods to facilitate the model's ability to refuse answering when appropriate. Our assumption is that image cropping enhances the model's visual perception of the specific question. By isolating the question, the model can more easily identify flaws or incompleteness in the visual input, triggering refusal explicitly.

\noindent\textbf{Experimental Setup:} 
We designed a comparative experiment with two input modes: 

\begin{itemize}
\addtolength{\leftskip}{-1em}
\item In Full-Page Mode, the complete and original test paper image serves as the input. Under this setting, the model exhibits three distinct behaviors: (1) generate an answer, (2) offer no response , or (3) provide an explicit refusal.

\item In Single-Question Mode, the specific unrecognizable questions are cropped from the page and input individually. In this scenario, model's behavior is categorized into two types: (1) generate an answer, or (2) provide an explicit refusal.
\end{itemize}

To evaluate the effectiveness, we categorize the results based on the joint behavior of the model in both modes. Let $N=110$ denote the total number of unrecognized questions. We define $n_{i}^{j}$ as the number of questions where the model exhibits behavior $i$ in Full-Page , $j$ in Single-Question, where $i$ represents Full-Page behaviors (1: Answer, 2: Silence, 3: Refusal) and $j$ denotes Single-Question behaviors (1: Answer, 2: Refusal).

Based on this notation, we propose three metrics to evaluate the performance:
\begin{itemize}
\addtolength{\leftskip}{-0.6em}
\item Refusal Failure Rate ($R_{fail}$): This metric measures the proportion of questions where the model persists in generating an answer:

\begin{equation}
    R_{fail} = \frac{\sum_{i=1}^{3} n_{i}^{1}}{N}
\end{equation}

\item Mitigation Rate ($R_{mitigation}$): This metric evaluates the effectiveness of the cropping strategy:
\begin{equation}
    R_{mitigation} = \frac{n_{1}^{2} + n_{2}^{2}}{N}
\end{equation}

\item Active Refusal Rate ($R_{active}$): This metric reflects the consistency of the model's refusal mechanism in both Modes.
\begin{equation}
    R_{active} = \frac{n_{3}^{2}}{N}
\end{equation}
\end{itemize}

\begin{figure}[t] 
\centering 
\includegraphics[width=1\linewidth]{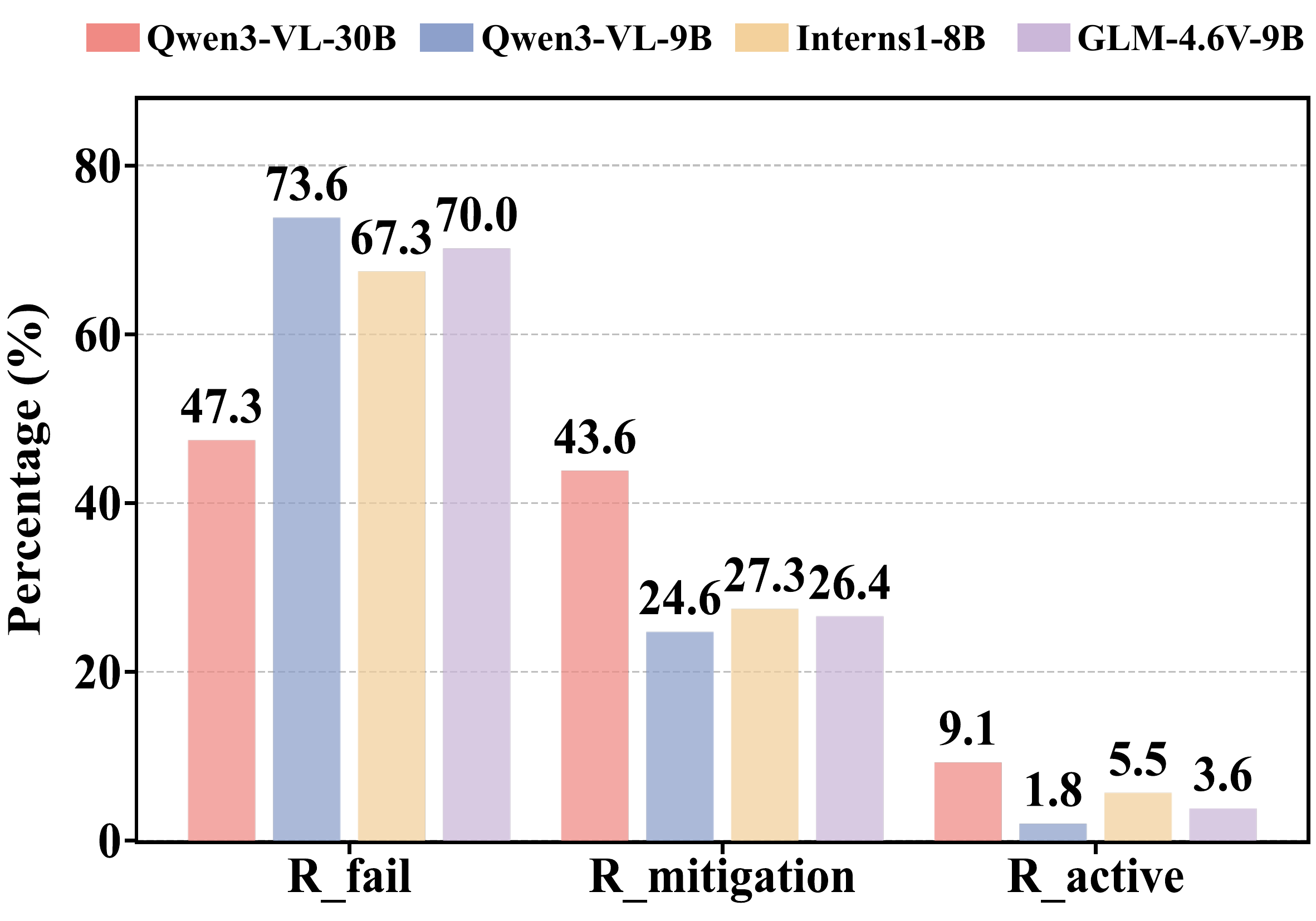} 
\caption{Refusal performance comparison under Full-Page and Single-Question input modes.} 
\label{fig:exper1} \end{figure}

\noindent\textbf{Result Analysis:}
As illustrated in Figure~\ref{fig:exper1}, in the full-page mode, models exhibit a significant deficiency in active refusal ($R_{active}$) (only 1.82\% for Qwen3vl-8b) alongside a high failure rate ($R_{fail}$), indicating a propensity to hallucinate responses rather than decline query. Conversely, upon applying the cropping strategy, we observe a substantial improvement in the models' ability to refuse incomplete inputs, as evidenced by the mitigation metrics ($R_{mitigation}$). These empirical results strictly validate our assumption: highlighting visual incompleteness through cropping effectively triggers the model's uncertainty, thereby enabling robust active refusal against incomplete visual contexts.

\section{Conclusion}
This paper addresses the lack of realistic benchmarks for noisy educational documents by introducing MathDoc, a dataset featuring authentic high school mathematics exams with diverse real-world artifacts and ``unrecognizable" samples, along with a comprehensive evaluation framework. MathDoc enables systematic assessments of both information extraction and active refusal capabilities, providing crucial insights for advancing model reliability. Its specific focus on refusal boundaries facilitates targeted model optimization, promoting more faithful and robust Large Language Models.

\section*{Limitations}
Though we initially explored visual perception enhancement strategies, we have not yet delved into model architecture design or pre-training objective optimization to fundamentally suppress the model's tendencies for speculative completion and forced transcription in incomplete contexts. Current evaluations serve more as a diagnostic tool rather than a generalizable algorithmic solution. Empowering models with the introspective capability to ``know what they do not know'' remains a critical challenge for building robust document processing systems.

\bibliographystyle{acl_natbib}
\bibliography{custom}

@article{meng2024deepstack,
  title={Deepstack: Deeply stacking visual tokens is surprisingly simple and effective for lmms},
  author={Meng, Lingchen and Yang, Jianwei and Tian, Rui and Dai, Xiyang and Wu, Zuxuan and Gao, Jianfeng and Jiang, Yu-Gang},
  journal={Advances in Neural Information Processing Systems},
  volume={37},
  pages={23464--23487},
  year={2024}
}

@article{bai2025qwen2,
  title={Qwen2. 5-vl technical report},
  author={Bai, Shuai and Chen, Keqin and Liu, Xuejing and Wang, Jialin and Ge, Wenbin and Song, Sibo and Dang, Kai and Wang, Peng and Wang, Shijie and Tang, Jun and others},
  journal={arXiv preprint arXiv:2502.13923},
  year={2025}
}

@article{fu2024ocrbench,
  title={Ocrbench v2: An improved benchmark for evaluating large multimodal models on visual text localization and reasoning},
  author={Fu, Ling and Kuang, Zhebin and Song, Jiajun and Huang, Mingxin and Yang, Biao and Li, Yuzhe and Zhu, Linghao and Luo, Qidi and Wang, Xinyu and Lu, Hao and others},
  journal={arXiv preprint arXiv:2501.00321},
  year={2024}
}

@inproceedings{ouyang2025omnidocbench,
  title={Omnidocbench: Benchmarking diverse pdf document parsing with comprehensive annotations},
  author={Ouyang, Linke and Qu, Yuan and Zhou, Hongbin and Zhu, Jiawei and Zhang, Rui and Lin, Qunshu and Wang, Bin and Zhao, Zhiyuan and Jiang, Man and Zhao, Xiaomeng and others},
  booktitle={Proceedings of the Computer Vision and Pattern Recognition Conference},
  pages={24838--24848},
  year={2025}
}

@article{lu2023mathvista,
  title={Mathvista: Evaluating mathematical reasoning of foundation models in visual contexts},
  author={Lu, Pan and Bansal, Hritik and Xia, Tony and Liu, Jiacheng and Li, Chunyuan and Hajishirzi, Hannaneh and Cheng, Hao and Chang, Kai-Wei and Galley, Michel and Gao, Jianfeng},
  journal={arXiv preprint arXiv:2310.02255},
  year={2023}
}

@inproceedings{zhang2024mathverse,
  title={Mathverse: Does your multi-modal llm truly see the diagrams in visual math problems?},
  author={Zhang, Renrui and Jiang, Dongzhi and Zhang, Yichi and Lin, Haokun and Guo, Ziyu and Qiu, Pengshuo and Zhou, Aojun and Lu, Pan and Chang, Kai-Wei and Qiao, Yu and others},
  booktitle={European Conference on Computer Vision},
  pages={169--186},
  year={2024},
  organization={Springer}
}

@article{feng2025mathreal,
  title={Mathreal: We keep it real! a real scene benchmark for evaluating math reasoning in multimodal large language models},
  author={Feng, Jun and Wang, Zixin and Zhang, Zhentao and Guo, Yue and Zhou, Zhihan and Chen, Xiuyi and Li, Zhenyang and Yin, Dawei},
  journal={arXiv preprint arXiv:2508.06009},
  year={2025}
}

@misc{bai2025qwen3vltechnicalreport,
      title={Qwen3-VL Technical Report}, 
      author={Shuai Bai and Yuxuan Cai and Ruizhe Chen and Keqin Chen and Xionghui Chen and Zesen Cheng and Lianghao Deng and Wei Ding and Chang Gao and Chunjiang Ge and Wenbin Ge and Zhifang Guo and Qidong Huang and Jie Huang and Fei Huang and Binyuan Hui and Shutong Jiang and Zhaohai Li and Mingsheng Li and Mei Li and Kaixin Li and Zicheng Lin and Junyang Lin and Xuejing Liu and Jiawei Liu and Chenglong Liu and Yang Liu and Dayiheng Liu and Shixuan Liu and Dunjie Lu and Ruilin Luo and Chenxu Lv and Rui Men and Lingchen Meng and Xuancheng Ren and Xingzhang Ren and Sibo Song and Yuchong Sun and Jun Tang and Jianhong Tu and Jianqiang Wan and Peng Wang and Pengfei Wang and Qiuyue Wang and Yuxuan Wang and Tianbao Xie and Yiheng Xu and Haiyang Xu and Jin Xu and Zhibo Yang and Mingkun Yang and Jianxin Yang and An Yang and Bowen Yu and Fei Zhang and Hang Zhang and Xi Zhang and Bo Zheng and Humen Zhong and Jingren Zhou and Fan Zhou and Jing Zhou and Yuanzhi Zhu and Ke Zhu},
      year={2025},
      eprint={2511.21631},
      archivePrefix={arXiv},
      primaryClass={cs.CV},
      url={https://arxiv.org/abs/2511.21631}, 
}

@misc{bai2025interns1scientificmultimodalfoundation,
      title={Intern-S1: A Scientific Multimodal Foundation Model}, 
      author={Lei Bai and Zhongrui Cai and Yuhang Cao and Maosong Cao and Weihan Cao and Chiyu Chen and Haojiong Chen and Kai Chen and Pengcheng Chen and Ying Chen and Yongkang Chen and Yu Cheng and Pei Chu and Tao Chu and Erfei Cui and Ganqu Cui and Long Cui and Ziyun Cui and Nianchen Deng and Ning Ding and Nanqing Dong and Peijie Dong and Shihan Dou and Sinan Du and Haodong Duan and Caihua Fan and Ben Gao and Changjiang Gao and Jianfei Gao and Songyang Gao and Yang Gao and Zhangwei Gao and Jiaye Ge and Qiming Ge and Lixin Gu and Yuzhe Gu and Aijia Guo and Qipeng Guo and Xu Guo and Conghui He and Junjun He and Yili Hong and Siyuan Hou and Caiyu Hu and Hanglei Hu and Jucheng Hu and Ming Hu and Zhouqi Hua and Haian Huang and Junhao Huang and Xu Huang and Zixian Huang and Zhe Jiang and Lingkai Kong and Linyang Li and Peiji Li and Pengze Li and Shuaibin Li and Tianbin Li and Wei Li and Yuqiang Li and Dahua Lin and Junyao Lin and Tianyi Lin and Zhishan Lin and Hongwei Liu and Jiangning Liu and Jiyao Liu and Junnan Liu and Kai Liu and Kaiwen Liu and Kuikun Liu and Shichun Liu and Shudong Liu and Wei Liu and Xinyao Liu and Yuhong Liu and Zhan Liu and Yinquan Lu and Haijun Lv and Hongxia Lv and Huijie Lv and Qitan Lv and Ying Lv and Chengqi Lyu and Chenglong Ma and Jianpeng Ma and Ren Ma and Runmin Ma and Runyuan Ma and Xinzhu Ma and Yichuan Ma and Zihan Ma and Sixuan Mi and Junzhi Ning and Wenchang Ning and Xinle Pang and Jiahui Peng and Runyu Peng and Yu Qiao and Jiantao Qiu and Xiaoye Qu and Yuan Qu and Yuchen Ren and Fukai Shang and Wenqi Shao and Junhao Shen and Shuaike Shen and Chunfeng Song and Demin Song and Diping Song and Chenlin Su and Weijie Su and Weigao Sun and Yu Sun and Qian Tan and Cheng Tang and Huanze Tang and Kexian Tang and Shixiang Tang and Jian Tong and Aoran Wang and Bin Wang and Dong Wang and Lintao Wang and Rui Wang and Weiyun Wang and Wenhai Wang and Jiaqi Wang and Yi Wang and Ziyi Wang and Ling-I Wu and Wen Wu and Yue Wu and Zijian Wu and Linchen Xiao and Shuhao Xing and Chao Xu and Huihui Xu and Jun Xu and Ruiliang Xu and Wanghan Xu and GanLin Yang and Yuming Yang and Haochen Ye and Jin Ye and Shenglong Ye and Jia Yu and Jiashuo Yu and Jing Yu and Fei Yuan and Yuhang Zang and Bo Zhang and Chao Zhang and Chen Zhang and Hongjie Zhang and Jin Zhang and Qiaosheng Zhang and Qiuyinzhe Zhang and Songyang Zhang and Taolin Zhang and Wenlong Zhang and Wenwei Zhang and Yechen Zhang and Ziyang Zhang and Haiteng Zhao and Qian Zhao and Xiangyu Zhao and Xiangyu Zhao and Bowen Zhou and Dongzhan Zhou and Peiheng Zhou and Yuhao Zhou and Yunhua Zhou and Dongsheng Zhu and Lin Zhu and Yicheng Zou},
      year={2025},
      eprint={2508.15763},
      archivePrefix={arXiv},
      primaryClass={cs.LG},
      url={https://arxiv.org/abs/2508.15763}, 
}

@article{wei2025deepseek,
  title={Deepseek-ocr: Contexts optical compression},
  author={Wei, Haoran and Sun, Yaofeng and Li, Yukun},
  journal={arXiv preprint arXiv:2510.18234},
  year={2025}
}

@article{hurst2024gpt,
  title={Gpt-4o system card},
  author={Hurst, Aaron and Lerer, Adam and Goucher, Adam P and Perelman, Adam and Ramesh, Aditya and Clark, Aidan and Ostrow, AJ and Welihinda, Akila and Hayes, Alan and Radford, Alec and others},
  journal={arXiv preprint arXiv:2410.21276},
  year={2024}
}

@article{comanici2025gemini,
  title={Gemini 2.5: Pushing the frontier with advanced reasoning, multimodality, long context, and next generation agentic capabilities},
  author={Comanici, Gheorghe and Bieber, Eric and Schaekermann, Mike and Pasupat, Ice and Sachdeva, Noveen and Dhillon, Inderjit and Blistein, Marcel and Ram, Ori and Zhang, Dan and Rosen, Evan and others},
  journal={arXiv preprint arXiv:2507.06261},
  year={2025}
}

@article{niu2025mineru2,
  title={Mineru2. 5: A decoupled vision-language model for efficient high-resolution document parsing},
  author={Niu, Junbo and Liu, Zheng and Gu, Zhuangcheng and Wang, Bin and Ouyang, Linke and Zhao, Zhiyuan and Chu, Tao and He, Tianyao and Wu, Fan and Zhang, Qintong and others},
  journal={arXiv preprint arXiv:2509.22186},
  year={2025}
}

@article{cui2025paddleocr,
  title={Paddleocr-vl: Boosting multilingual document parsing via a 0.9 b ultra-compact vision-language model},
  author={Cui, Cheng and Sun, Ting and Liang, Suyin and Gao, Tingquan and Zhang, Zelun and Liu, Jiaxuan and Wang, Xueqing and Zhou, Changda and Liu, Hongen and Lin, Manhui and others},
  journal={arXiv preprint arXiv:2510.14528},
  year={2025}
}

@inproceedings{yang2025cc,
  title={Cc-ocr: A comprehensive and challenging ocr benchmark for evaluating large multimodal models in literacy},
  author={Yang, Zhibo and Tang, Jun and Li, Zhaohai and Wang, Pengfei and Wan, Jianqiang and Zhong, Humen and Liu, Xuejing and Yang, Mingkun and Wang, Peng and Bai, Shuai and others},
  booktitle={Proceedings of the IEEE/CVF International Conference on Computer Vision},
  pages={21744--21754},
  year={2025}
}

@inproceedings{heakl2025kitab,
  title={Kitab-bench: A comprehensive multi-domain benchmark for arabic ocr and document understanding},
  author={Heakl, Ahmed and Sohail, Muhammad Abdullah and Ranjan, Mukul and Elbadry, Rania and Ahmad, Ghazi Shazan and El-Geish, Mohamed and Maher, Omar and Shen, Zhiqiang and Khan, Fahad Shahbaz and Khan, Salman},
  booktitle={Findings of the Association for Computational Linguistics: ACL 2025},
  pages={22006--22024},
  year={2025}
}

@article{chen2025mint,
  title={MINT-CoT: Enabling Interleaved Visual Tokens in Mathematical Chain-of-Thought Reasoning},
  author={Chen, Xinyan and Zhang, Renrui and Jiang, Dongzhi and Zhou, Aojun and Yan, Shilin and Lin, Weifeng and Li, Hongsheng},
  journal={arXiv preprint arXiv:2506.05331},
  year={2025}
}

@article{ye2025logicocr,
  title={LogicOCR: Do Your Large Multimodal Models Excel at Logical Reasoning on Text-Rich Images?},
  author={Ye, Maoyuan and Zhang, Jing and Liu, Juhua and Du, Bo and Tao, Dacheng},
  journal={arXiv preprint arXiv:2505.12307},
  year={2025}
}

@article{li2020docbank,
  title={Docbank: A benchmark dataset for document layout analysis},
  author={Li, Minghao and Xu, Yiheng and Cui, Lei and Huang, Shaohan and Wei, Furu and Li, Zhoujun and Zhou, Ming},
  journal={arXiv preprint arXiv:2006.01038},
  year={2020}
}

@inproceedings{zhou2025dogr,
  title={DOGR: Towards Versatile Visual Document Grounding and Referring},
  author={Zhou, Yinan and Chen, Yuxin and Lin, Haokun and Wu, Yichen and Yang, Shuyu and Qi, Zhongang and Ma, Chen and Zhu, Li},
  booktitle={Proceedings of the IEEE/CVF International Conference on Computer Vision},
  pages={3596--3606},
  year={2025}
}

@article{liu2025deepseek,
  title={Deepseek-v3. 2: Pushing the frontier of open large language models},
  author={Liu, Aixin and Mei, Aoxue and Lin, Bangcai and Xue, Bing and Wang, Bingxuan and Xu, Bingzheng and Wu, Bochao and Zhang, Bowei and Lin, Chaofan and Dong, Chen and others},
  journal={arXiv preprint arXiv:2512.02556},
  year={2025}
}

@article{ma2024mmlongbench,
  title={Mmlongbench-doc: Benchmarking long-context document understanding with visualizations},
  author={Ma, Yubo and Zang, Yuhang and Chen, Liangyu and Chen, Meiqi and Jiao, Yizhu and Li, Xinze and Lu, Xinyuan and Liu, Ziyu and Ma, Yan and Dong, Xiaoyi and others},
  journal={Advances in Neural Information Processing Systems},
  volume={37},
  pages={95963--96010},
  year={2024}
}

@inproceedings{mathew2021docvqa,
  title={Docvqa: A dataset for vqa on document images},
  author={Mathew, Minesh and Karatzas, Dimosthenis and Jawahar, CV},
  booktitle={Proceedings of the IEEE/CVF winter conference on applications of computer vision},
  pages={2200--2209},
  year={2021}
}

@article{horn2025benchmarking,
  title={Benchmarking Document Parsers on Mathematical Formula Extraction from PDFs},
  author={Horn, Pius and Keuper, Janis},
  journal={arXiv preprint arXiv:2512.09874},
  year={2025}
}

@article{bai2025complex,
  title={Complex Mathematical Expression Recognition: Benchmark, Large-Scale Dataset and Strong Baseline},
  author={Bai, Weikang and Du, Yongkun and Su, Yuchen and Xie, Yazhen and Chen, Zhineng},
  journal={arXiv preprint arXiv:2512.13731},
  year={2025}
}

@article{liu2024ocrbench,
  title={Ocrbench: on the hidden mystery of ocr in large multimodal models},
  author={Liu, Yuliang and Li, Zhang and Huang, Mingxin and Yang, Biao and Yu, Wenwen and Li, Chunyuan and Yin, Xu-Cheng and Liu, Cheng-Lin and Jin, Lianwen and Bai, Xiang},
  journal={Science China Information Sciences},
  volume={67},
  number={12},
  pages={220102},
  year={2024},
  publisher={Springer}
}

@article{li2024llava,
  title={Llava-onevision: Easy visual task transfer},
  author={Li, Bo and Zhang, Yuanhan and Guo, Dong and Zhang, Renrui and Li, Feng and Zhang, Hao and Zhang, Kaichen and Zhang, Peiyuan and Li, Yanwei and Liu, Ziwei and others},
  journal={arXiv preprint arXiv:2408.03326},
  year={2024}
}

@article{chen2025advancing,
  title={Advancing mathematical reasoning in language models: The impact of problem-solving data, data synthesis methods, and training stages},
  author={Chen, Zui and Liu, Tianqiao and Tian, Mi and Tong, Qing and Luo, Weiqi and Liu, Zitao},
  journal={arXiv preprint arXiv:2501.14002},
  year={2025}
}

@inproceedings{liu2025cmm,
  title={Cmm-math: A chinese multimodal math dataset to evaluate and enhance the mathematics reasoning of large multimodal models},
  author={Liu, Wentao and Pan, Qianjun and Zhang, Yi and Liu, Zhuo and Wu, Ji and Zhou, Jie and Zhou, Aimin and Chen, Qin and Jiang, Bo and He, Liang},
  booktitle={Proceedings of the 33rd ACM International Conference on Multimedia},
  pages={12585--12591},
  year={2025}
}

@article{blecher2023nougat,
  title={Nougat: Neural optical understanding for academic documents},
  author={Blecher, Lukas and Cucurull, Guillem and Scialom, Thomas and Stojnic, Robert},
  journal={arXiv preprint arXiv:2308.13418},
  year={2023}
}

@inproceedings{zhong2020image,
  title={Image-based table recognition: data, model, and evaluation},
  author={Zhong, Xu and ShafieiBavani, Elaheh and Jimeno Yepes, Antonio},
  booktitle={European conference on computer vision},
  pages={564--580},
  year={2020},
  organization={Springer}
}

@inproceedings{rasheed2024glamm,
  title={Glamm: Pixel grounding large multimodal model},
  author={Rasheed, Hanoona and Maaz, Muhammad and Shaji, Sahal and Shaker, Abdelrahman and Khan, Salman and Cholakkal, Hisham and Anwer, Rao M and Xing, Eric and Yang, Ming-Hsuan and Khan, Fahad S},
  booktitle={Proceedings of the IEEE/CVF Conference on Computer Vision and Pattern Recognition},
  pages={13009--13018},
  year={2024}
}

@article{zhu2025internvl3,
  title={Internvl3: Exploring advanced training and test-time recipes for open-source multimodal models},
  author={Zhu, Jinguo and Wang, Weiyun and Chen, Zhe and Liu, Zhaoyang and Ye, Shenglong and Gu, Lixin and Tian, Hao and Duan, Yuchen and Su, Weijie and Shao, Jie and others},
  journal={arXiv preprint arXiv:2504.10479},
  year={2025}
}

@inproceedings{zhong2019publaynet,
  title={Publaynet: largest dataset ever for document layout analysis},
  author={Zhong, Xu and Tang, Jianbin and Yepes, Antonio Jimeno},
  booktitle={2019 International conference on document analysis and recognition (ICDAR)},
  pages={1015--1022},
  year={2019},
  organization={IEEE}
}

@inproceedings{jaume2019funsd,
  title={Funsd: A dataset for form understanding in noisy scanned documents},
  author={Jaume, Guillaume and Ekenel, Hazim Kemal and Thiran, Jean-Philippe},
  booktitle={2019 International Conference on Document Analysis and Recognition Workshops (ICDARW)},
  volume={2},
  pages={1--6},
  year={2019},
  organization={IEEE}
}

@inproceedings{hu2024mplug,
  title={mplug-docowl 1.5: Unified structure learning for ocr-free document understanding},
  author={Hu, Anwen and Xu, Haiyang and Ye, Jiabo and Yan, Ming and Zhang, Liang and Zhang, Bo and Zhang, Ji and Jin, Qin and Huang, Fei and Zhou, Jingren},
  booktitle={Findings of the Association for Computational Linguistics: EMNLP 2024},
  pages={3096--3120},
  year={2024}
}

@inproceedings{kim2019textbook,
  title={Textbook question answering with multi-modal context graph understanding and self-supervised open-set comprehension},
  author={Kim, Daesik and Kim, Seonhoon and Kwak, Nojun},
  booktitle={Proceedings of the 57th Annual Meeting of the Association for Computational Linguistics},
  pages={3568--3584},
  year={2019}
}

@article{chen2024humans,
  title={Humans or llms as the judge? a study on judgement biases},
  author={Chen, Guiming Hardy and Chen, Shunian and Liu, Ziche and Jiang, Feng and Wang, Benyou},
  journal={arXiv preprint arXiv:2402.10669},
  year={2024}
}

@inproceedings{li2020improving,
  title={Improving attention-based handwritten mathematical expression recognition with scale augmentation and drop attention},
  author={Li, Zhe and Jin, Lianwen and Lai, Songxuan and Zhu, Yecheng},
  booktitle={2020 17th International Conference on Frontiers in Handwriting Recognition (ICFHR)},
  pages={175--180},
  year={2020},
  organization={IEEE}
}

@article{yan2025mathagent,
  title={Mathagent: Leveraging a mixture-of-math-agent framework for real-world multimodal mathematical error detection},
  author={Yan, Yibo and Wang, Shen and Huo, Jiahao and Yu, Philip S and Hu, Xuming and Wen, Qingsong},
  journal={arXiv preprint arXiv:2503.18132},
  year={2025}
}

@article{hariyanto2025artificial,
  title={Artificial intelligence in adaptive education: a systematic review of techniques for personalized learning},
  author={Hariyanto and Kristianingsih, Francisca Xaveria Diah and Maharani, Rizqona},
  journal={Discover Education},
  volume={4},
  number={1},
  pages={458},
  year={2025},
  publisher={Springer}
}

@inproceedings{wang2025wilddoc,
  title={Wilddoc: How far are we from achieving comprehensive and robust document understanding in the wild?},
  author={Wang, An-Lan and Tang, Jingqun and Liao, Lei and Feng, Hao and Liu, Qi and Fei, Xiang and Lu, Jinghui and Wang, Han and Liu, Hao and Liu, Yuliang and others},
  booktitle={Proceedings of the 2025 Conference on Empirical Methods in Natural Language Processing},
  pages={23002--23012},
  year={2025}
}

\clearpage
\appendix
\section{Appendix}
\label{sec:appendix}
\subsection{The example of dataset}
\begin{figure*}[t]
\centering
\begin{tabular}{@{}m{0.48\textwidth}@{\hspace{0.04\textwidth}}m{0.48\textwidth}@{}}
\centering
\includegraphics[width=\linewidth]{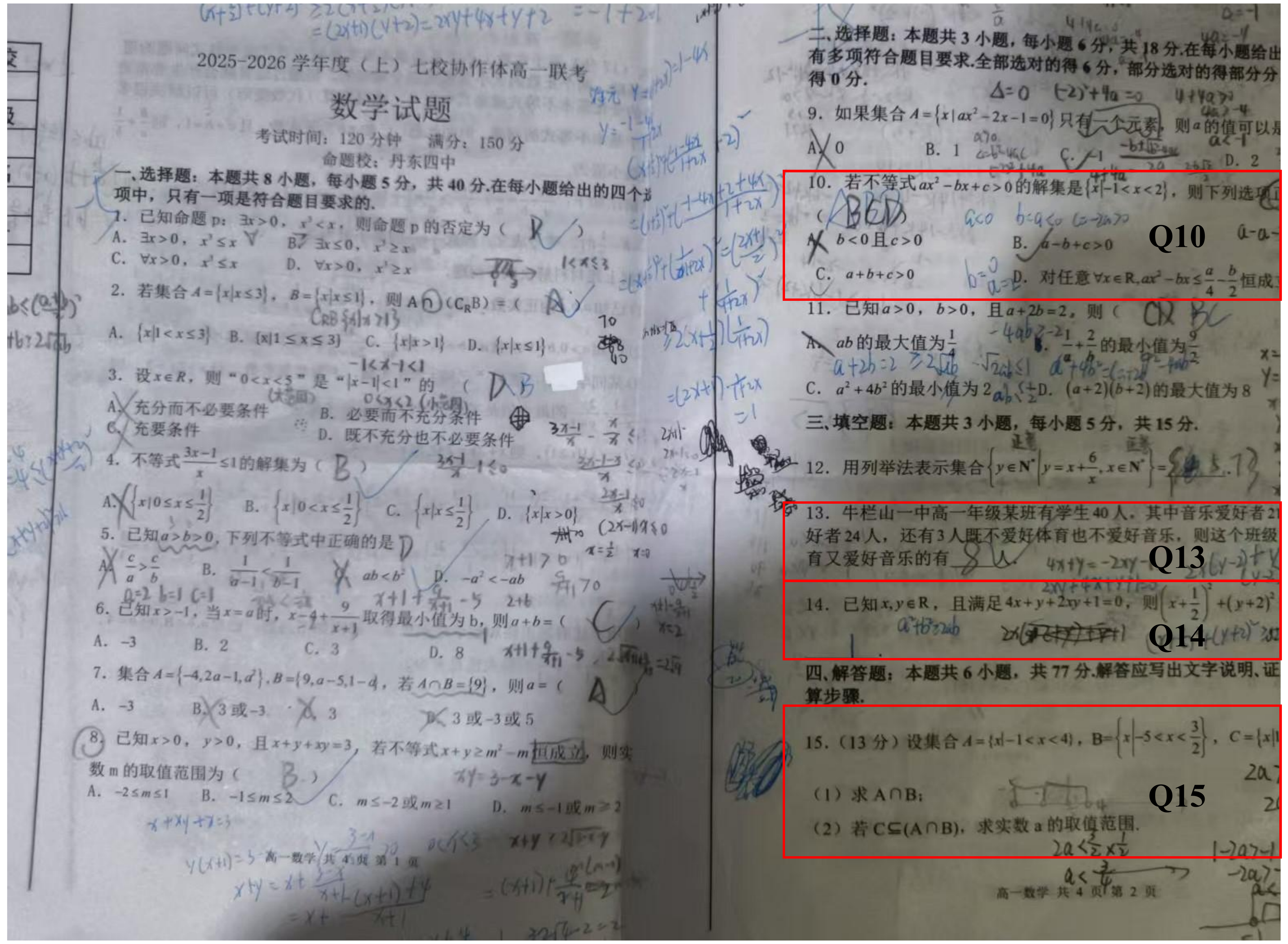}
\captionof*{figure}{(a)}
\vspace{2mm}
\includegraphics[width=\linewidth]{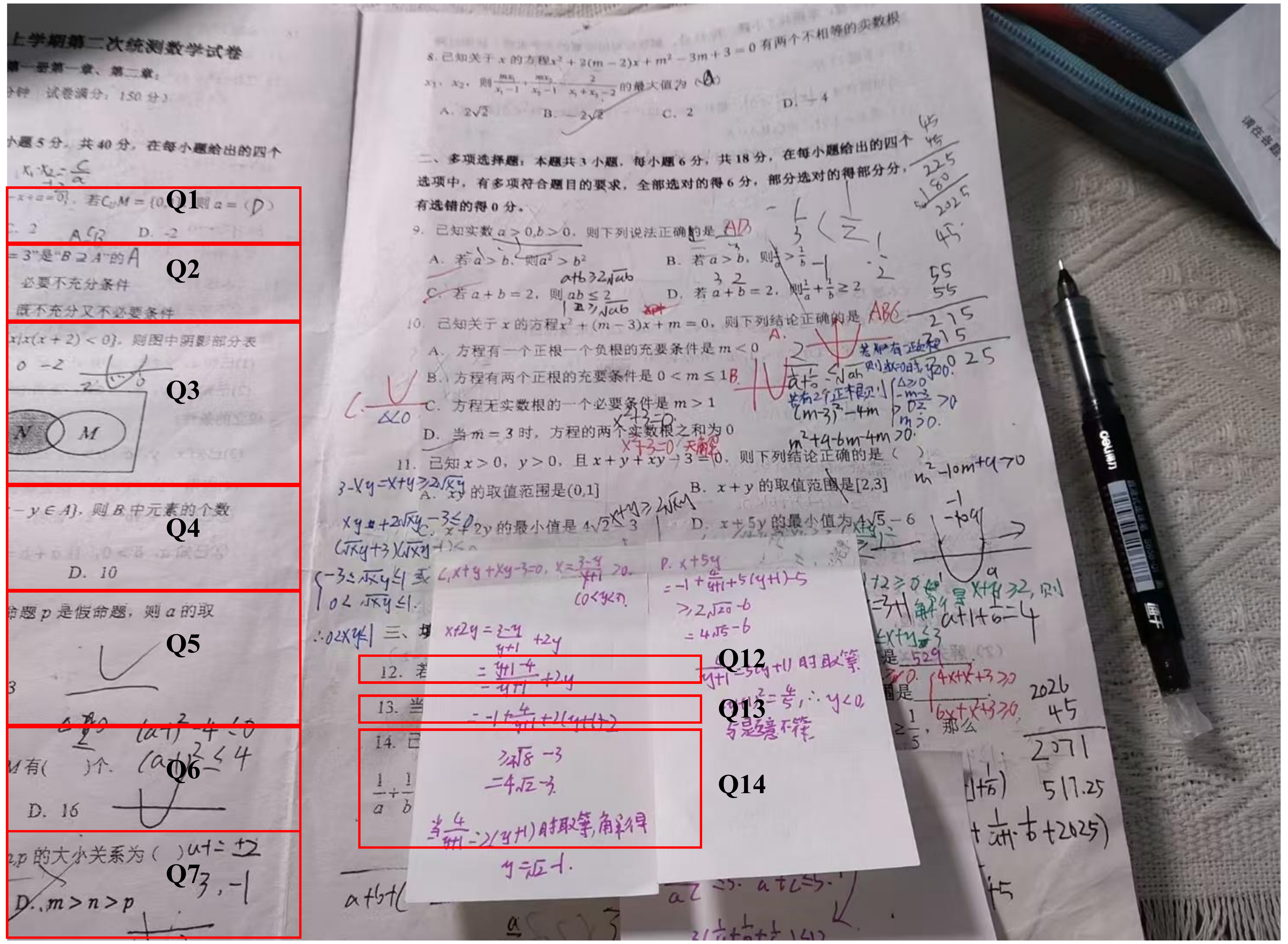}
\captionof*{figure}{(b)}
&
\centering
\includegraphics[width=\linewidth]{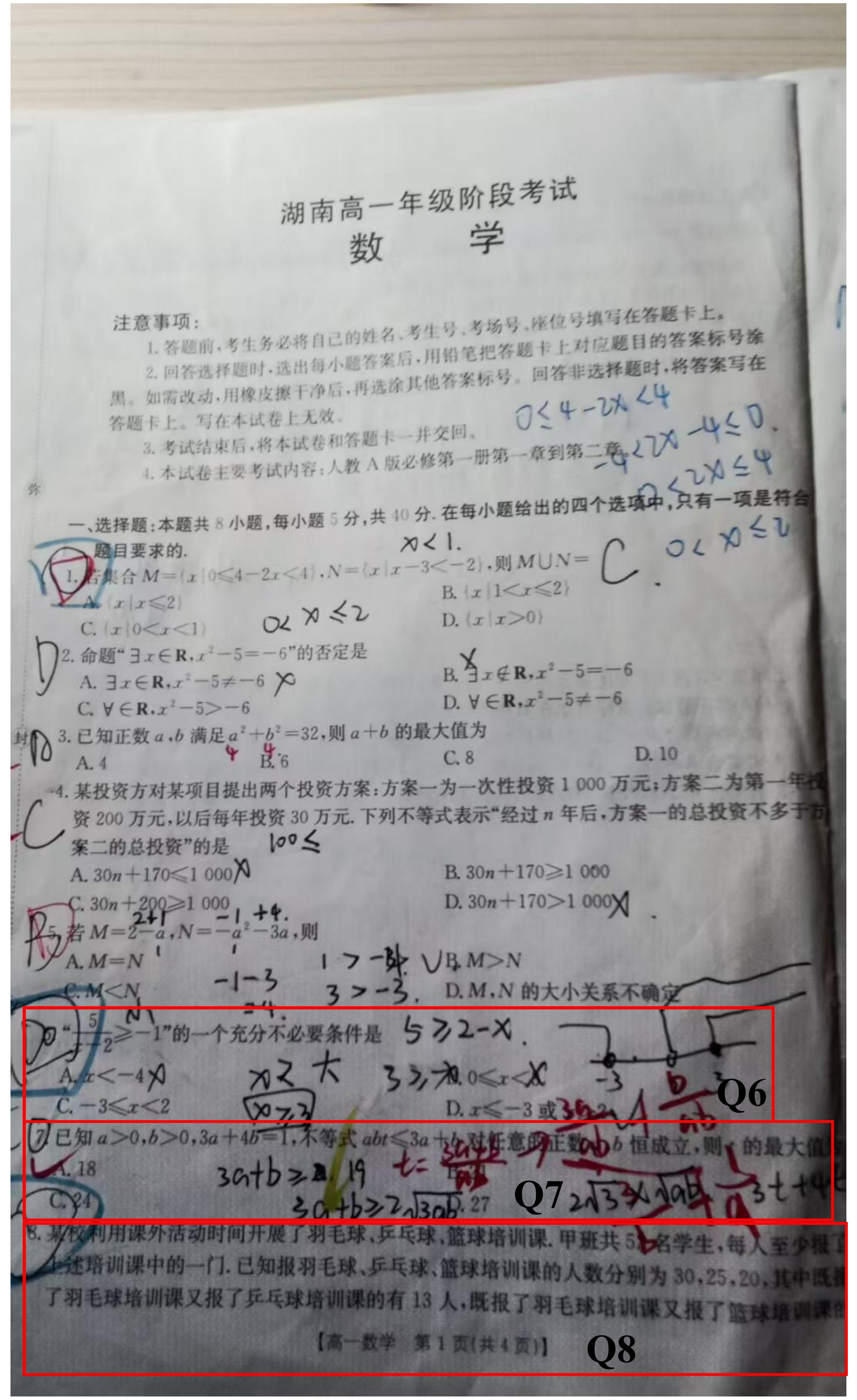}
\captionof*{figure}{(c)}
\end{tabular}
\caption{Examples of three categories of unrecognizable cases defined in our work:
(i) heavy handwriting occlusions and corrections in critical text regions;
(ii) physical obstructions (e.g., stains and wrinkles); and
(iii) compromised document integrity, such as incomplete questions or folded pages.
Unrecognizable question regions are highlighted with red bounding boxes, as these degradations remove critical semantic information and violate the legibility constraints, thereby marked as [unrecognizable].}
\label{fig:appendix-unrecog}
\end{figure*}

In this section, we present representative examples from the MathDoc dataset, illustrating unrecognizable questions in Figure~\ref{fig:appendix-unrecog}.

\subsection{The fairness of LLM as judge}
We note that our evaluation relies on Large Language Model as automatic judge. Rather than claiming absolute objectivity, we aim to demonstrate that under our constrained evaluation protocol, the judge exhibits consistent and interpretable behavior.

\begin{figure*}[t]
  \centering
  \includegraphics[width=\textwidth]{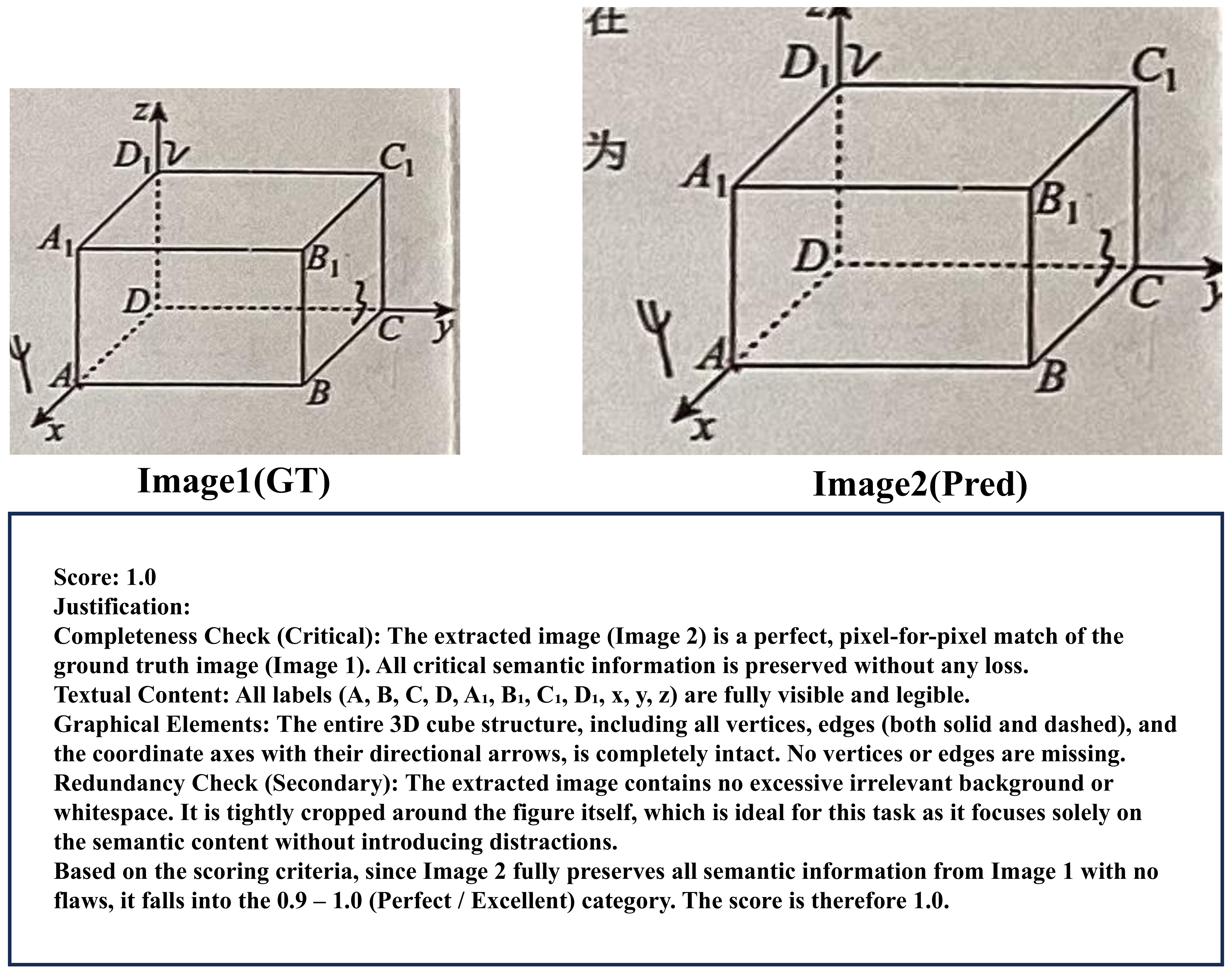}
  \caption{An example of LLM-based image semantic similarity evaluation}
  \label{fig:whynotiou}
\end{figure*}

\begin{figure*}[t]
  \centering
  \includegraphics[width=\textwidth]{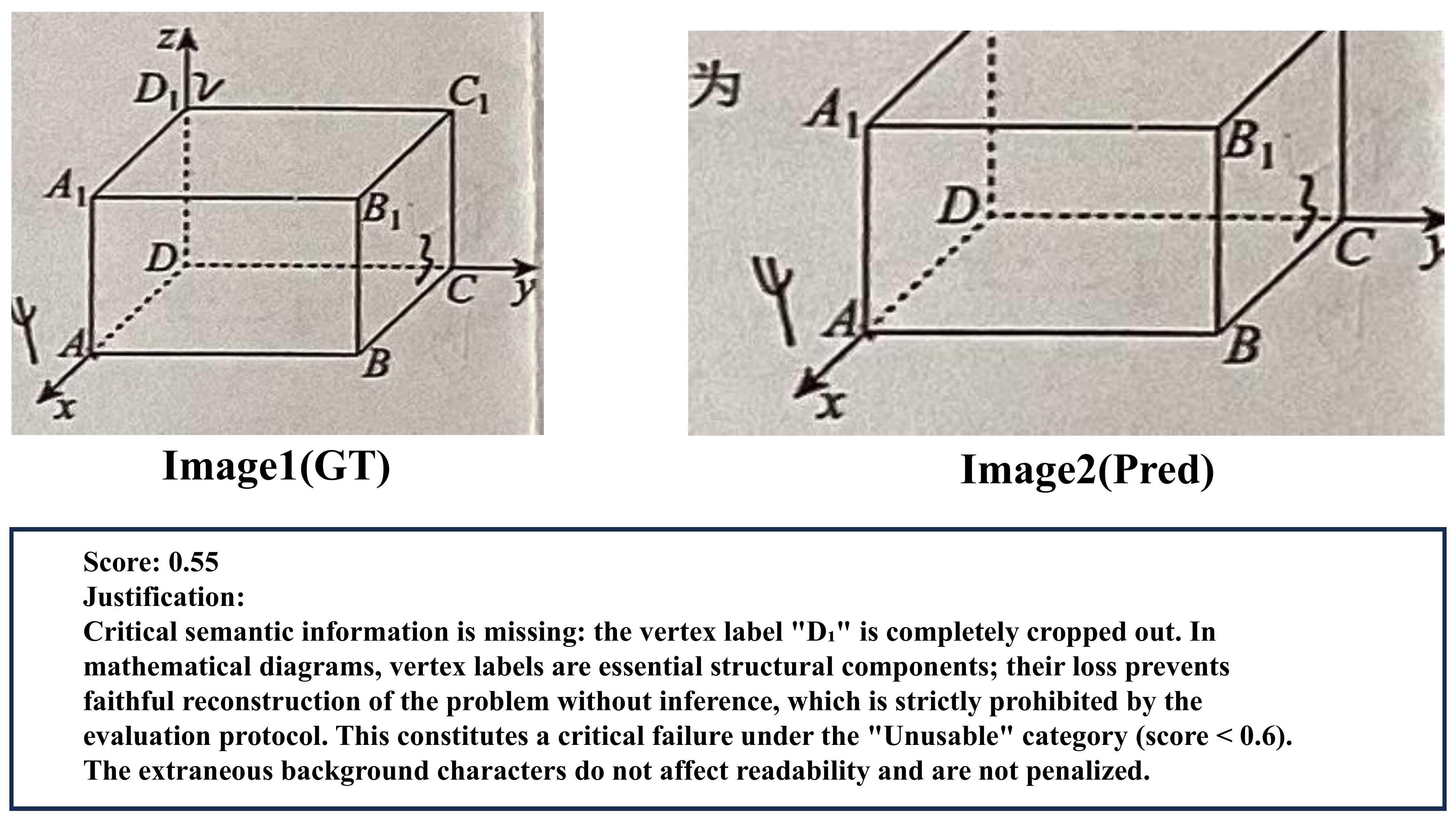}
  \caption{A failure case correctly identified by the LLM-based semantic judge}
  \label{fig:imagejudger}
\end{figure*}

First, for image-based verification, the judge is restricted to assessing semantic consistency under a fixed instruction,
rather than subjective visual quality.
We therefore avoid pixel-level metrics such as IoU, which are highly sensitive to common layout variations in real-world documents and may not reliably reflect semantic validity.
As shown in Figure~\ref{fig:whynotiou}, two images with moderate IoU can still be semantically equivalent and are correctly assigned a high score by the LLM-based judge.
In contrast, Figure~\ref{fig:imagejudger} presents a visually similar prediction that omits critical semantic elements and is accordingly penalized.

\begin{figure*}[t]
  \centering
  \includegraphics[width=\textwidth]{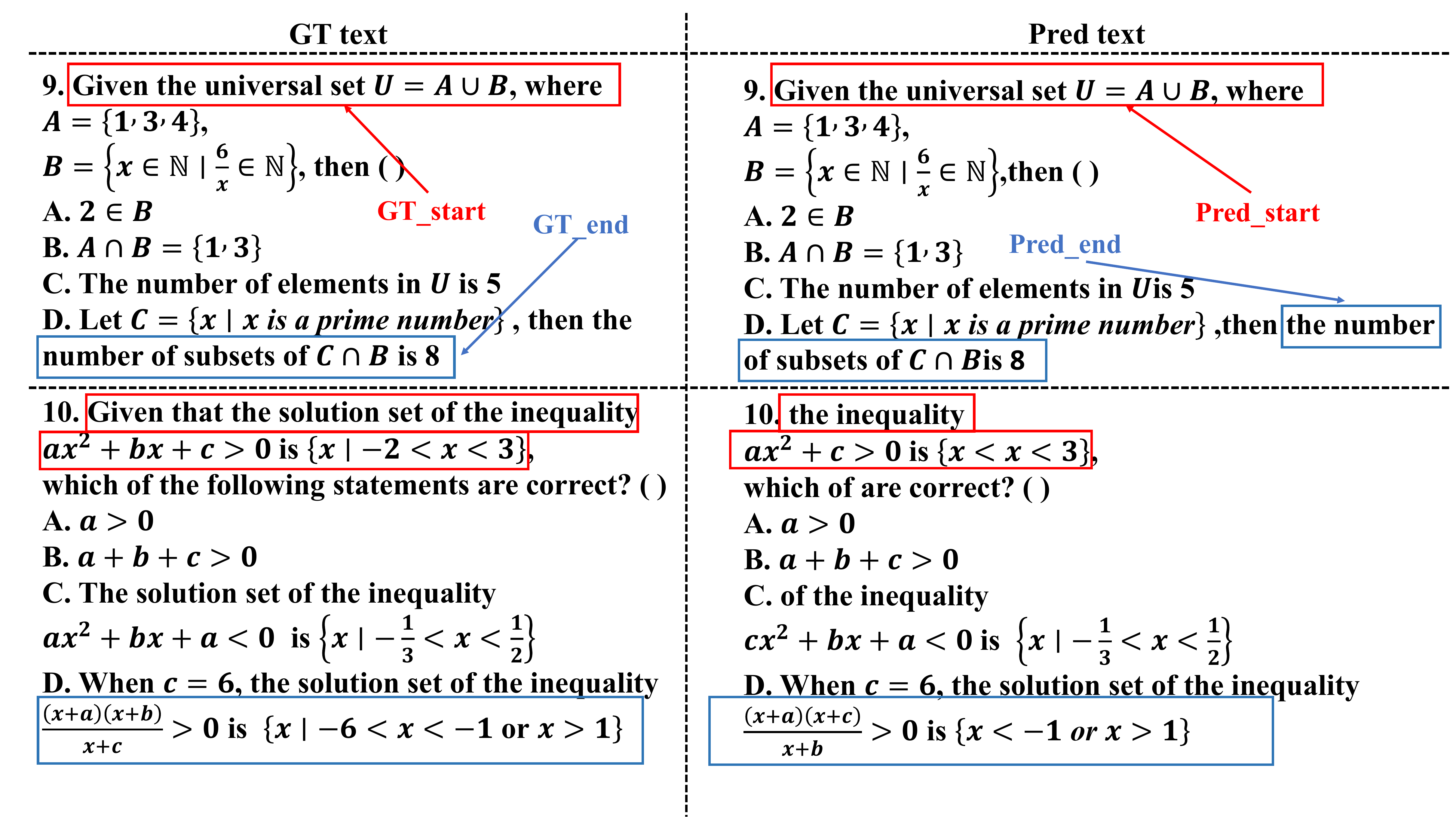}
  \caption{Illustration of dual-anchor semantic alignment between ground-truth and predicted text. The left hand-side two questions are all from GT, the right hand-side two are all from Pred text stream.}
  \label{fig:Align}
\end{figure*}

Second, at the text level, the judge is designed to tolerate common OCR-induced noise while preserving strict semantic alignment between the predicted and ground-truth questions. Figure~\ref{fig:Align} provides a representative matching example.

\subsection{Prompt Details}
In this section, we detail the prompts used in our evaluation pipeline. For readability, we present the English version of the prompt. The experiments are conducted using the original Chinese prompt, which is semantically equivalent. Specifically, we describe: (i) the prompt used to guide a language model in identifying the start and end indices of the predicted text spans within the generated output (Figure~\ref{fig:ocr_alignment_prompt}); (ii) the prompt employed in an \textbf{MLLM-as-a-judge} setting to assess semantic similarity between ground-truth figures and model-predicted cropped images (Figure~\ref{fig:visual_completeness_prompt}); and (iii) the prompt provided to \textbf{MLLMs} for document-level information extraction (Figure~\ref{fig:structure_transcription_prompt}).

\vspace{-0.8em} 

\begin{figure*}[t]
\centering
  \begin{tcolorbox}[
    colback=white,
    colframe=black,
    boxrule=0.4pt,
    arc=1mm,
    left=2mm,
    right=2mm,
    top=1mm,
    bottom=1mm,
    fontupper=\footnotesize
  ]

  \textbf{Role.} You are a \emph{Visual Semantic Completeness Evaluator} with expertise in
  the structured assessment of mathematical documents.
  Your objective is to evaluate the semantic completeness of an
  \emph{Extracted Image (Image~2)} relative to its corresponding
  \emph{Ground Truth Image (Image~1)}.

  \vspace{0.5em}
  \textbf{Task.} Given a pair of images, assess whether Image~2 faithfully preserves
  all semantic information present in Image~1.
  This includes textual content, mathematical expressions, and graphical elements.

  \vspace{0.6em}
  \textbf{1. Evaluation Procedure}

  \begin{itemize}[leftmargin=*, itemsep=0.2em]
    \item \textbf{Completeness Check (Critical).}
    Carefully compare Image~2 against Image~1 and determine whether any
    critical semantic information is missing.
    This includes, but is not limited to:
    truncated text, missing vertices or edges in geometric figures,
    and blurred or omitted superscripts/subscripts in formulas.

    \item \textbf{Redundancy Check (Secondary).}
    Examine whether Image~2 contains excessive irrelevant background
    or whitespace.
    Additional whitespace is acceptable as long as it does not interfere
    with content readability or interpretation.
  \end{itemize}

  \vspace{0.6em}
  \textbf{2. Scoring Criteria}

  \begin{itemize}[leftmargin=*, itemsep=0.2em]
    \item \textbf{0.9 -- 1.0 (Perfect / Excellent).}
    Image~2 fully preserves all semantic information from Image~1.

    \item \textbf{0.6 -- 0.8 (Pass / Minor Flaws).}
    The semantic content is fundamentally complete with negligible defects.

    \item \textbf{0.0 -- 0.59 (Unusable).}
    Any semantic or structural loss that prevents full reconstruction
    of the original problem meaning.
  \end{itemize}

  \end{tcolorbox}

  \caption{
 Prompt used for evaluating the visual similarity of extracted mathematical document images.
  }
  \label{fig:visual_completeness_prompt}
\end{figure*}

\begin{figure*}[t]
\centering
\begin{tcolorbox}[
  colback=white,
  colframe=black,
  boxrule=0.4pt,
  arc=1mm,
  left=2mm,
  right=2mm,
  top=1mm,
  bottom=1mm,
  fontupper=\footnotesize
]

\textbf{Role.} You are an \emph{OCR Text Alignment Expert}.  
Your objective is to map structured questions from the \emph{Ground Truth (GT)} to their
\emph{physical boundaries} within the noisy \emph{Predicted Text (Pred)}.

\vspace{0.5em}
\textbf{Task.} Establish a correspondence between clean GT questions and noisy Pred content.
For each GT question, identify the exact \texttt{pred\_start\_snippet} and
\texttt{pred\_end\_snippet} in Pred.

\vspace{0.6em}
\textbf{1. Core Principles}

\begin{itemize}[leftmargin=*, itemsep=0.2em]
  \item \textbf{Monotonicity (Strict Order).}
  Question order in Pred must strictly follow GT.
  The start of Question $N$ must appear \emph{after} the end of Question $N-1$.
  No backtracking or overlap is allowed.

  \item \textbf{Verbatim Extraction.}
  \texttt{pred\_start\_snippet} and \texttt{pred\_end\_snippet} must be copied \emph{exactly}
  from Pred, preserving all OCR errors, garbled characters, and missing punctuation.
  Do \emph{not} normalize or correct the text.

  \item \textbf{Physical Truncation Strategy.}
  Do not struggle to search for a semantic ending of Question $N$.
  Instead, if the end can not be found, try to locate the start of Question $N{+}1$.
  The end of Question $N$ is defined as the text segment immediately preceding it.

  \item \textbf{Robustness to Failure.}
  If Question $N$ cannot be matched due to severe noise or missing content,
  return empty strings and \emph{immediately proceed} to align Question $N{+}1$.
  Maximize the number of aligned questions.
\end{itemize}

\vspace{0.6em}
\textbf{2. Alignment Strategy}

For each GT question:

\begin{itemize}[leftmargin=*, itemsep=0.2em]
  \item \textbf{Locate Start.}
  Identify the question beginning in Pred using \emph{semantic similarity}.
  Prioritize topic-level matching (e.g., \emph{Complex Numbers}, \emph{Vectors})
  even when mathematical expressions are distorted
  (e.g., $z=1+i$ vs.\ $z=2/(1+i)$).

  \item \textbf{Ignore Status Tags.}
  Skip metadata or preprocessing tags (e.g., \texttt{[Format Normal]})
  and begin extraction from the actual question content or index.

  \item \textbf{Locate End.}
  Identify the start of Question $N{+}1$.
  The end of Question $N$ is the text segment strictly before it.
  Garbled text is acceptable as long as it lies within the physical boundary.
\end{itemize}

\vspace{0.6em}
\textbf{3. Output Schema}

Return a \emph{pure JSON list} containing objects with the following fields:

\begin{center}
\texttt{\{question\_id, gt\_start\_snippet, gt\_end\_snippet,}
\texttt{pred\_start\_snippet, pred\_end\_snippet, pred\_answer\}}
\end{center}

\vspace{0.4em}
Extract answers as concise values (e.g., \texttt{A}, \texttt{3}, \texttt{[Answer: X]}).
Return empty strings if no answer is found.
Do \emph{not} extract full question text.

\vspace{0.6em}
\textbf{4. Few-Shot}

\begin{itemize}[leftmargin=*, itemsep=0.2em]
  \item \textbf{Distorted Mathematics.}
  GT: \texttt{z=2/(1+i)}; Pred: \texttt{z=1+i}.
  Match based on the shared semantic topic.

  \item \textbf{Missing Index.}
  GT: \texttt{18. Function ...}; Pred: \texttt{(17pts) Function ...}.
  Match using semantic context despite index loss.

  \item \textbf{Unrecognizable Content.}
  If the GT marks a question as \texttt{[Unrecognizable]}, inspect the corresponding section in \texttt{Pred}. Consider the question **successfully aligned** if the \texttt{Pred} content is either:
  1) \textbf{Empty} (null string), or
  2) Explicitly contains \textbf{unrecognizability indicators} (e.g., ``unrecognizable", ``illegible", or failure tokens).
  In these cases, extract the empty string or the specific indicator as the snippet and proceed to the next question.
\end{itemize}

\end{tcolorbox}
\caption{Full prompt used for OCR question-to-text boundary alignment under severe noise.}
\label{fig:ocr_alignment_prompt}
\end{figure*}

\begin{figure*}[t]
\centering
\begin{tcolorbox}[
  colback=white,
  colframe=black,
  boxrule=0.4pt,
  arc=1mm,
  left=2mm,
  right=2mm,
  top=1mm,
  bottom=1mm,
  fontupper=\footnotesize
]

\textbf{Role.} You are a \emph{Strict Document Structure Analysis Specialist}.
Your objective is to accurately transcribe the printed structure of exam papers
without performing mathematical problem-solving or speculative completion.

\vspace{0.6em}
\textbf{Global Constraints:} 
\textbf{(1) No Subjective Solving:}
  Do not solve mathematical problems or reason about correctness.
\textbf{(2) No Hallucination:}
  If printed text is occluded or missing, do not guess or complete it.
\textbf{(3) Strict Mode Switching:}
  Apply transcription rules strictly according to the question type
  defined below.

\vspace{0.6em}
\textbf{Workflow:} Execute the following logic pipeline sequentially for each detected question region.

\vspace{0.4em}
\textbf{Step 1: Legibility \& Integrity Check.}

\emph{Condition:}
If any critical printed information (question index, stem content,
or specific options) is blurred, occluded, or missing due to folding,
stains, or damage.

\emph{Action:}
Immediately output:
\begin{center}
\texttt{[ID]. [Unrecognizable]}
\end{center}
and terminate processing for this question.
Do not attempt to repair or infer the text.

\vspace{0.6em}
\textbf{Step 2: Question Mode Classification \& Transcription}

Classify the question into one of the following modes and apply the
corresponding strict rules.

\vspace{0.4em}
\textbf{Mode A: Subjective Questions (Calculation / Proof).}

\emph{Characteristics:}
High point value (e.g., $>$10 points), sub-questions such as (1)(2),
or directives like ``Prove'' or ``Solve''.
\emph{Rules:} Transcribe the printed question stem only. Strictly ignore all handwritten content.
\vspace{0.4em}

\textbf{Mode B: Objective Questions (Multiple Choice / Fill-in-the-Blank).}
\emph{Characteristics:}
Presence of options (A/B/C/D) or underline placeholders.

\emph{Rules:}
\begin{itemize}[leftmargin=*, itemsep=0.15em]
  \item Transcribe all printed options (A, B, C, D) completely.
  \item \textbf{Answer Extraction:} If a handwritten mark (e.g., a tick or letter) is explicitly visible, append \texttt{[Answer: X]} at the end. If the answer area is blank, output printed text only.
  \item \textbf{Prohibition:}
  Do not infer answers based on internal knowledge.
  If no handwriting is present, do not generate an answer tag.
\end{itemize}

\vspace{0.6em}
\textbf{Step 3: Figure Extraction}

\emph{Target:}
Computer-generated diagrams only (e.g., geometric figures or function plots).

\emph{Action:}
Append an image marker after the question stem using normalized coordinates:
\begin{center}
\texttt{<!-- Image (x1, y1, x2, y2) -->}
\end{center}
where coordinates are normalized to the range $[0,1000]$.

\vspace{0.6em}
\textbf{Few-Shot Demonstrations (Illustrative)}

\textbf{Example 1: Objective (Answered).}

\emph{Input:}
Question 5 with handwritten ``C'' marked in the answer bracket.

\emph{Output:}
\begin{quote}
Let $M=\{1,2\}, N=\{2,3\}$, then $M \cup N =$ ( ) \texttt{[Answer: C]} \\
A. $\{1\}$ \quad
B. $\{2\}$ \quad
C. $\{1,2,3\}$ \quad
D. $\{3\}$
\end{quote}

\vspace{0.4em}
\textbf{Example 2: Objective (Unanswered).}

\emph{Input:}
Question 6 with an empty answer bracket.

\emph{Output:}
\begin{quote}
Which of the following functions is an even function? ( ) \\
A. $y=x$ \quad
B. $y=x^2$ \quad
C. $y=\sin x$ \quad
D. $y=e^x$
\end{quote}
(\emph{No answer tag is generated.})

\vspace{0.4em}
\textbf{Example 3: Subjective (Handwriting Ignored).}

\emph{Input:}
Question 17 (10 points) with handwritten formulas covering whitespace.

\emph{Output:}
\begin{quote}
(10 points) As shown in the figure... \\
(1) Prove that... \\
\texttt{<!-- Image (50, 60, 150, 160) -->}
\end{quote}

\vspace{0.4em}
\textbf{Example 4: Illegible Input (Refusal).}

\emph{Input:}
Question 8 partially covered by a page fold.

\emph{Output:}
\begin{quote}
\texttt{[Unrecognizable]}
\end{quote}

\end{tcolorbox}
\caption{Prompt specifying strict rules for document structure analysis and transcription of exam papers.}
\label{fig:structure_transcription_prompt}
\end{figure*}

\end{document}